\documentclass[11pt, a4paper, logo, copyright]{googledeepmind}

\usepackage[authoryear, sort&compress, round]{natbib}
\usepackage{multirow}
\usepackage{tcolorbox}
\usepackage{graphicx} 
\usepackage{subcaption}
\usepackage{float}
\usepackage{cleveref}
\usepackage[parfill]{parskip}
\usepackage{pifont}

\newcommand\subsubsubsection[1]{\textbf{#1}}
\usepackage{textcomp}
\usepackage{afterpage}
\usepackage{makecell}
\usepackage{xcolor,colortbl}
\hypersetup{
    colorlinks=true,
}

\title{General Geospatial Inference with a Population Dynamics Foundation Model}

\correspondingauthor{$\{$gautamprasad, davidsch$\}$@google.com}

\reportnumber{} 

\renewcommand{\today}{2025-01-29}

\author[$\circ$,1]{Mohit Agarwal}
\author[$\circ$,1]{Mimi Sun}
\author[1]{Chaitanya Kamath}
\author[1]{Arbaaz Muslim}
\author[1,2]{Prithul Sarker}
\author[1]{Joydeep Paul}
\author[1]{\\Hector Yee}
\author[1]{Marcin Sieniek}
\author[1]{Kim Jablonski}
\author[1]{Swapnil Vispute}
\author[1]{Atul Kumar}
\author[1]{Yael Mayer}
\author[1]{David Fork}
\author[1]{Sheila de Guia}
\author[1]{Jamie McPike}
\author[1]{Adam Boulanger}
\author[1]{Tomer Shekel}
\author[1]{David Schottlander}
\author[1]{Yao Xiao}
\author[1]{\\Manjit Chakravarthy Manukonda}
\author[1]{Yun Liu}
\author[1]{Neslihan Bulut}
\author[1]{Sami Abu-el-haija}
\author[1]{Bryan Perozzi}
\author[1]{\\Monica Bharel}
\author[1]{Von Nguyen}
\author[1]{Luke Barrington}
\author[1]{Niv Efron}
\author[1]{Yossi Matias}
\author[1]{Greg Corrado}
\author[3]{Krish Eswaran}
\author[1]{\\Shruthi Prabhakara}
\author[1]{Shravya Shetty}
\author[1]{Gautam Prasad}

\affil[$\circ$]{equal contribution}
\affil[1]{Google Research}
\affil[2]{University of Nevada, Reno}
\affil[3]{PURI AI}

\begin{abstract}
Supporting the health and well-being of dynamic populations around the world requires governmental agencies, organizations, and researchers to understand and reason over complex relationships between human behavior and local contexts. This support includes identifying populations at elevated risk and gauging where to target limited aid resources. Traditional approaches to these classes of problems often entail developing manually curated, task-specific features and models to represent human behavior and the natural and built environment, which can be challenging to adapt to new, or even related tasks. To address this, we introduce the Population Dynamics Foundation Model (PDFM), which aims to capture the relationships between diverse data modalities and is applicable to a broad range of geospatial tasks. We first construct a geo-indexed dataset for postal codes and counties across the United States, capturing rich aggregated information on human behavior from maps, busyness, and aggregated search trends, and environmental factors such as weather and air quality. We then model this data and the complex relationships between locations using a graph neural network, producing embeddings that can be adapted to a wide range of downstream tasks using relatively simple models. We evaluate the effectiveness of our approach by benchmarking it on 27 downstream tasks spanning three distinct domains: health indicators, socioeconomic factors, and environmental measurements. The approach achieves state-of-the-art performance on geospatial interpolation across all tasks, surpassing existing satellite and geotagged image based location encoders. In addition, it achieves state-of-the-art performance in extrapolation and super-resolution for 25 of the 27 tasks. We also show that the PDFM can be combined with a state-of-the-art forecasting foundation model, TimesFM, to predict unemployment and poverty, achieving performance that surpasses fully supervised forecasting. The full set of embeddings and sample code are publicly available for researchers. In conclusion, we have demonstrated a general purpose approach to geospatial modeling tasks critical to understanding population dynamics by leveraging a rich set of complementary globally available datasets that can be readily adapted to previously unseen machine learning tasks.
\end{abstract}

\begin{document}

\maketitle

\section{Introduction}\label{intro}

Models of population dynamics can be a powerful tool in understanding the impact of environmental, social, and economic factors on our well-being.  Greater population level data could contribute to better planning and resource allocation, leading to improved outcomes for some of public health’s most vexing problems including noncommunicable disease, behavioral health disorders and climate related health impacts. There is evidence that having a deeper understanding of your postal code is a better predictor of long term health than your genetics code (\cite{graham2016your}).

Whether we’re modeling the impact of an opioid epidemic or the population migration after an environmental disaster, such problems have traditionally needed dedicated teams and datasets to understand and model a single target variable (\cite{gupta2003uses, monnat2019using}). To scale the capabilities and accessibility of these types of geospatial models, we introduce a novel Population Dynamics Foundation Model (PDFM) that uses machine learning to synthesize rich, globally available geospatial data such as maps, busyness, and aggregated search trends, along with environmental signals such as weather and air quality. To simplify the modeling of a broad range of problems affecting populations around the world, our approach allows these datasets to be at variable resolutions, consistency, and quality. For example, aggregated search trends might be insightful in heavily populated regions, while maps could be more useful in areas with lower internet penetration. We also show that incorporating embeddings from PDFM can imbue existing models with geospatial reasoning capabilities.

\begin{figure} 
    \centering
    \includegraphics[width=\textwidth]{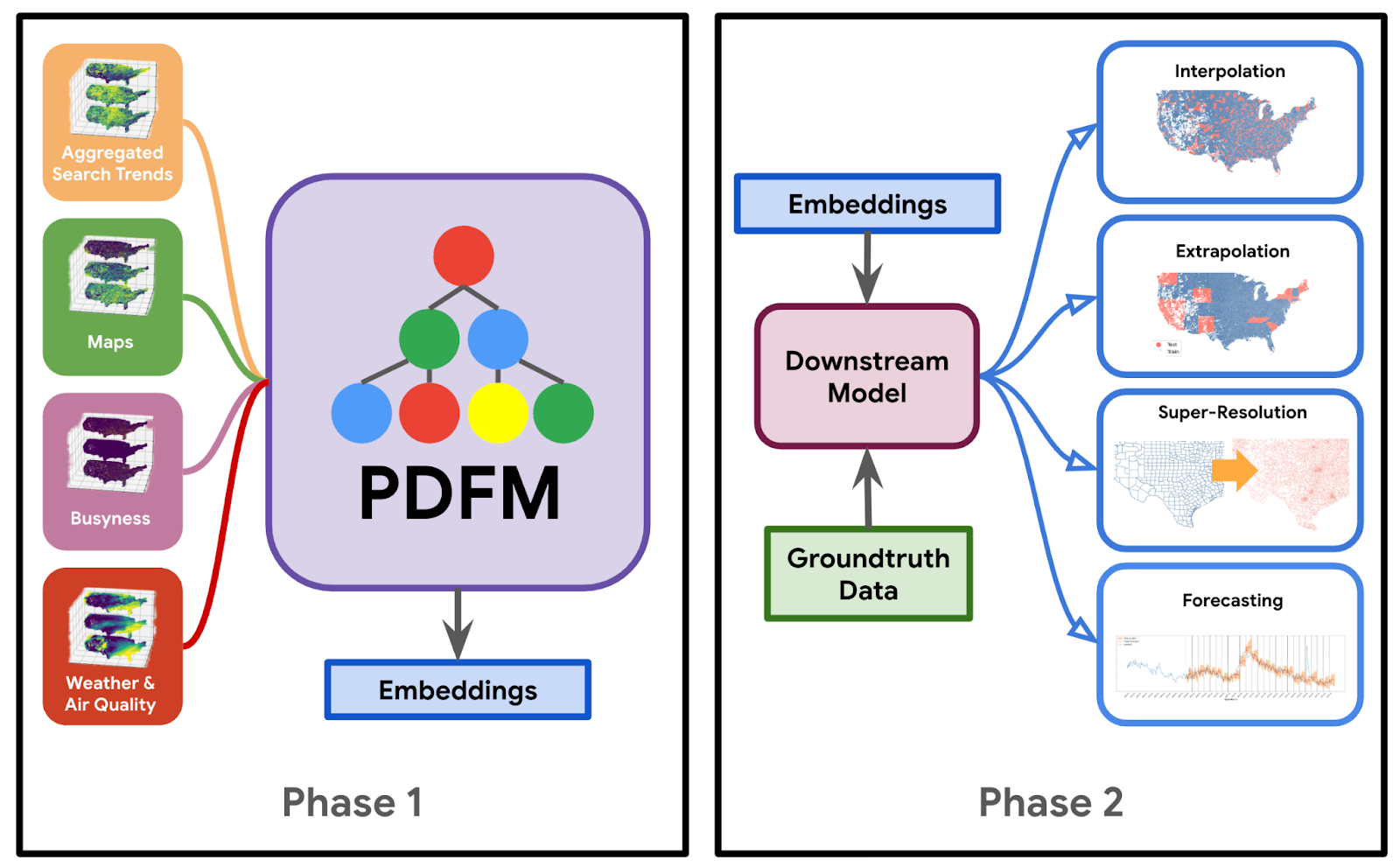}
    \caption{\footnotesize\textbf{Training and Applying the Population Dynamics Foundation Model.} Visual summary of the breakdown of phases to solve geospatial problems using our model. In Phase 1 we combine datasets with our graph neural network (GNN) architecture to train a foundation model that produces embeddings that are generally useful and not tied to a specific task. In Phase 2, the embeddings and existing task specific groundtruth are used to learn a downstream model (such as linear regression, simple multilayer perceptrons, or gradient boosted decision trees) that can be applied to a range of tasks including interpolation, extrapolation, super-resolution, and forecasting.}
    \label{fig:phase_diagram}
\end{figure}

\begin{figure} 
    \centering
    \includegraphics[width=\textwidth]{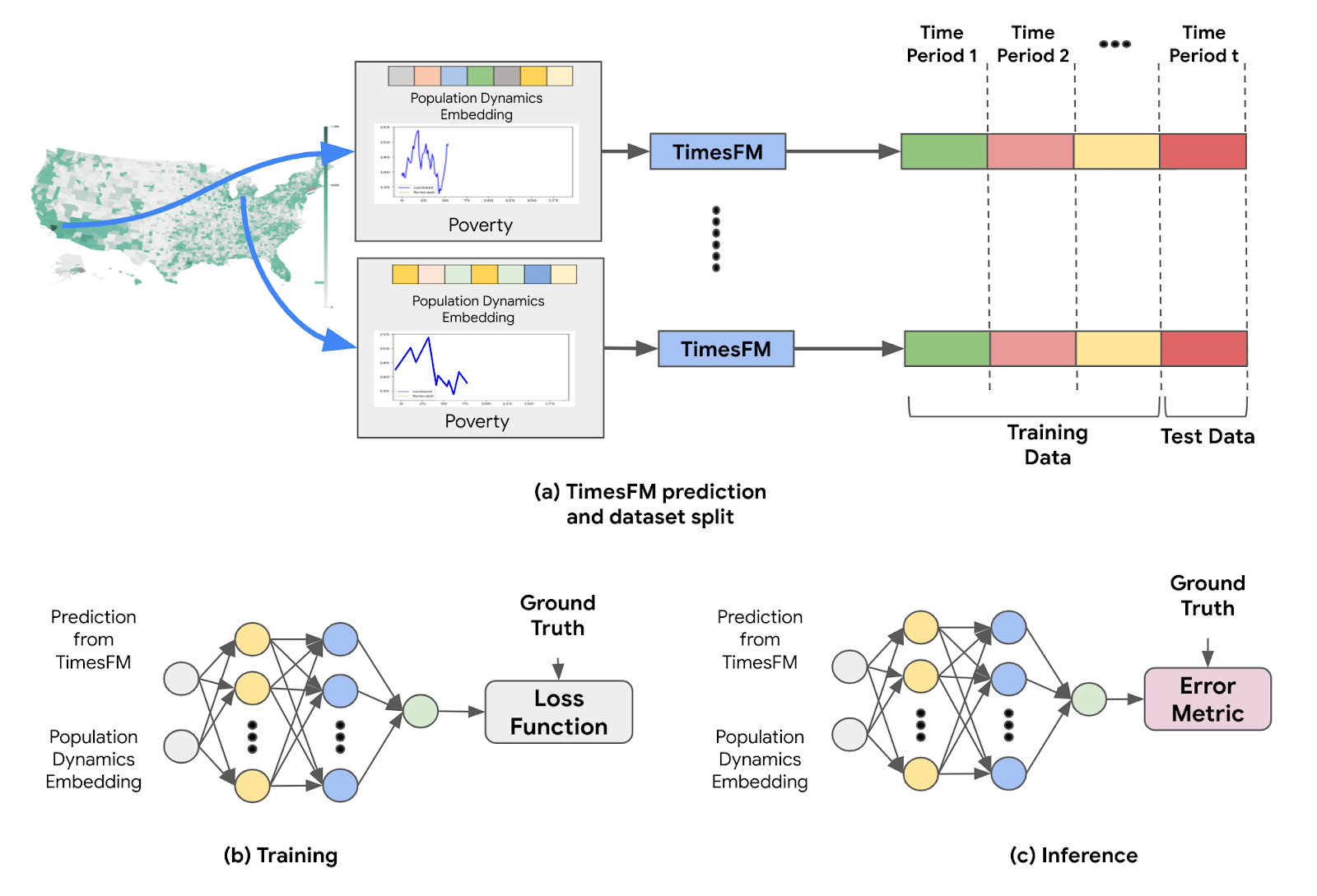}
    \caption{\footnotesize\textbf{Downstream Application to Existing Forecasting Model.} Illustration of forecasting pipeline with PDFM embedding and TimesFM prediction. (a) shows how TimesFM predictions are integrated into the training and test dataset creation process. (b) and (c) demonstrate the training and inference procedure of the multilayer perceptron (MLP) respectively.}
    \label{fig:timesfm_architecture}
\end{figure}

PDFM’s architecture leverages a graph neural network (GNN) because of its inherent ability to embed the network relationships across locations and handle missing data via message passing. We show that embeddings derived from the GNN are generally useful across a wide range of downstream tasks in spatial interpolation, extrapolation, and super-resolution problems. For these tasks, we show that a variety of relatively simple downstream models including linear regression, simple neural networks, and gradient boosted decision trees (GBDTs) all achieve similar performance when using PDFM-synthesized embeddings.

We refer to the combination of our unique datasets (maps, busyness, aggregated search trends, weather, and air quality) and the GNN architecture as “the PDFM” throughout this paper and specifically note cases with additional input data (the PDFM with SatCLIP). The presented experiments focus on the contiguous United States as illustrative examples; generalization to other regions and tasks is possible.

Our core contributions are as follows.
\begin{itemize}
\item \textbf{Population Dynamics Foundation Model.} We introduce a flexible foundation model framework that encodes and compresses diverse datasets (we evaluate maps, busyness, aggregated search trends, weather, air quality, and remote sensing) at different spatial resolutions to solve geospatial problems efficiently and intuitively using GNNs. See Figure \ref{fig:phase_diagram} for a visualization of our input and evaluation signals.
\item \textbf{SoTA Performance on Health, Socioeconomic, and Environment Geospatial Tasks.} We benchmark the PDFM on interpolation, extrapolation, and super-resolution problems across 27 tasks (\cite{sun2024community}) covering health, socioeconomics, and the environment. Our foundation model provides SoTA performance in all 27 tasks for interpolation and 25 tasks for both extrapolation and super-resolution. 
\item \textbf{An Understanding of the Complementarity of Individual Datasets.} We introduce a disentangled embedding architecture that partitions the embedding dimensions by data source, ensuring that the model attends to all inputs and retains pertinent information from each, while also providing data source-level interpretability for downstream tasks. For example, individually the best signal for tree cover in our model comes from weather and air quality, while maps provide the strongest signal for high cholesterol rates, and aggregated web search is the best at identifying chronic obstructive pulmonary disease (COPD) rates. We show that the PDFM is able to significantly exceed individual dataset performance by combining and enhancing the utility of these datasets.
\item \textbf{Downstream Application to Forecasting Model.} We show how the PDFM augments the SoTA forecasting foundation model TimesFM (\cite{das2023decoder}), to improve forecasts like unemployment at the county level and poverty at the postal code level (Figure \ref{fig:timesfm_architecture}). To accomplish this, we train a simple error correction model using PDFM embeddings and the TimesFM forecast for a single timestep. The resulting model surpasses a fully supervised forecasting method without fine-tuning any of the foundation models. Similar methods can be used to augment other existing geospatial classification and regression models with the PDFM embeddings.
\item \textbf{Surpassing of Satellite Imagery and Geotagged Image Based Embeddings.}Our experiments show that the PDFM embeddings surpass both the SatCLIP by \cite{klemmer2023satclip} and GeoCLIP by \cite{cepeda2024geoclip} embeddings in every downstream task except for tree cover extrapolation and elevation super-resolution. Previous results show a tradeoff where GeoCLIP outperformed satellite based approaches for human centric applications while PDFM is able to distill both human centric and built/natural environment signals.
\item \textbf{Practical Utility of the PDFM in Research, Social Good, Public \& Environmental Health, and Business.}Through the strong performance on downstream interpolation, extrapolation, super-resolution and forecasting tasks, we demonstrate that the PDFM can easily be extended to a range of applications that require geospatial modeling capabilities in research, social good, health, and business. 
\item \textbf{Public Release of Embeddings for Researchers.}We make all of the embeddings and sample code available on GitHub to apply PDFM to novel use cases and empower the research community. 
\end{itemize}

\section{Previous Work}\label{previous_work}

Using machine learning approaches that leverage a variety of data sources has been shown to expand the spatiotemporal resolution and coverage of populations represented in geospatial maps. \cite{deville2014dynamic} used a linear regression model with population-weighted least squares to study population movement over time using mobile phone data in Portugal and France. Web search trends data has been used to model automobile sales, unemployment claims, travel destination planning, and consumer confidence in \cite{choi2012predicting} for countries around the world using linear models. Finely crafted web search features have been used in a regression model to predict influenza at the state level in \cite{ginsberg2009detecting}. \cite{rolf2021generalizable} used a featurization method over satellite imagery followed by task-specific regression to model forest cover, elevation, population density, nighttime lights, road length, and housing prices. A long short-term memory (LSTM) network was trained on precipitation, temperature, solar radiation, thermal radiation, snowfall, surface pressure, and basin attributes to forecast floods in \cite{nearing2024global}. \cite{lam2023learning} used a graph neural network to forecast weather using historical data globally. Each of these techniques provide unique abilities to model downstream geospatial tasks with strong performance that is actionable for policy makers and governments. However, each also required hand crafted features derived from each data source (mobile phone usage, web search, or satellite imagery) and in most cases, a custom model for each use case. This limits the ability to scale these types of models to new use cases and also increases the complexity of utilizing complementary datasets (for example, if we wanted to use both web search and mobile phone usage data).

There have also been a variety of previous techniques that have sought to build a general geographic location encoding or geographic foundation model. Most focus on specific domains such as internet data, remote sensing imagery, or map databases. A semantic embedding for geo-coordinates was used to build GPS2Vec by \cite{yin2019gps2vec} using geotagged internet data including check-ins, tweets, and images. Satellite Contrastive Location-Image Pretraining (SatCLIP) \cite{klemmer2023satclip} is a “global, general-purpose geographic location encoder” based on two years of satellite imagery. SatCLIP outperforms a previous version called MOSAIKS (\cite{rolf2021generalizable}) and a technique using contrastive self-supervision, CSP by \cite{mai2023csp}. GeoCLIP by \cite{cepeda2024geoclip} is an image-to-GPS model using a CLIP-inspired alignment between GPS and images from 4.72 million Flickr images. GeoCLIP often outperforms SatCLIP. We benchmarked our foundation model against both GeoCLIP and SatCLIP. A fine-tuned version of an LLM using OpenStreetMap by \cite{manvi2023geollm} was used to predict things such as population density, income, and home value. They focused on how performance varied with different language models and used a night lights dataset with GBDTs as a baseline. Our approach builds on this work by providing a wider range of general purpose datasets complementing human behavior signals (like those in GeoCLIP) with built and natural environment signals (like those in SatCLIP).

\section{Experimental Results}\label{experimental_results}

\begin{figure} 
    \centering
    \includegraphics[width=\textwidth]{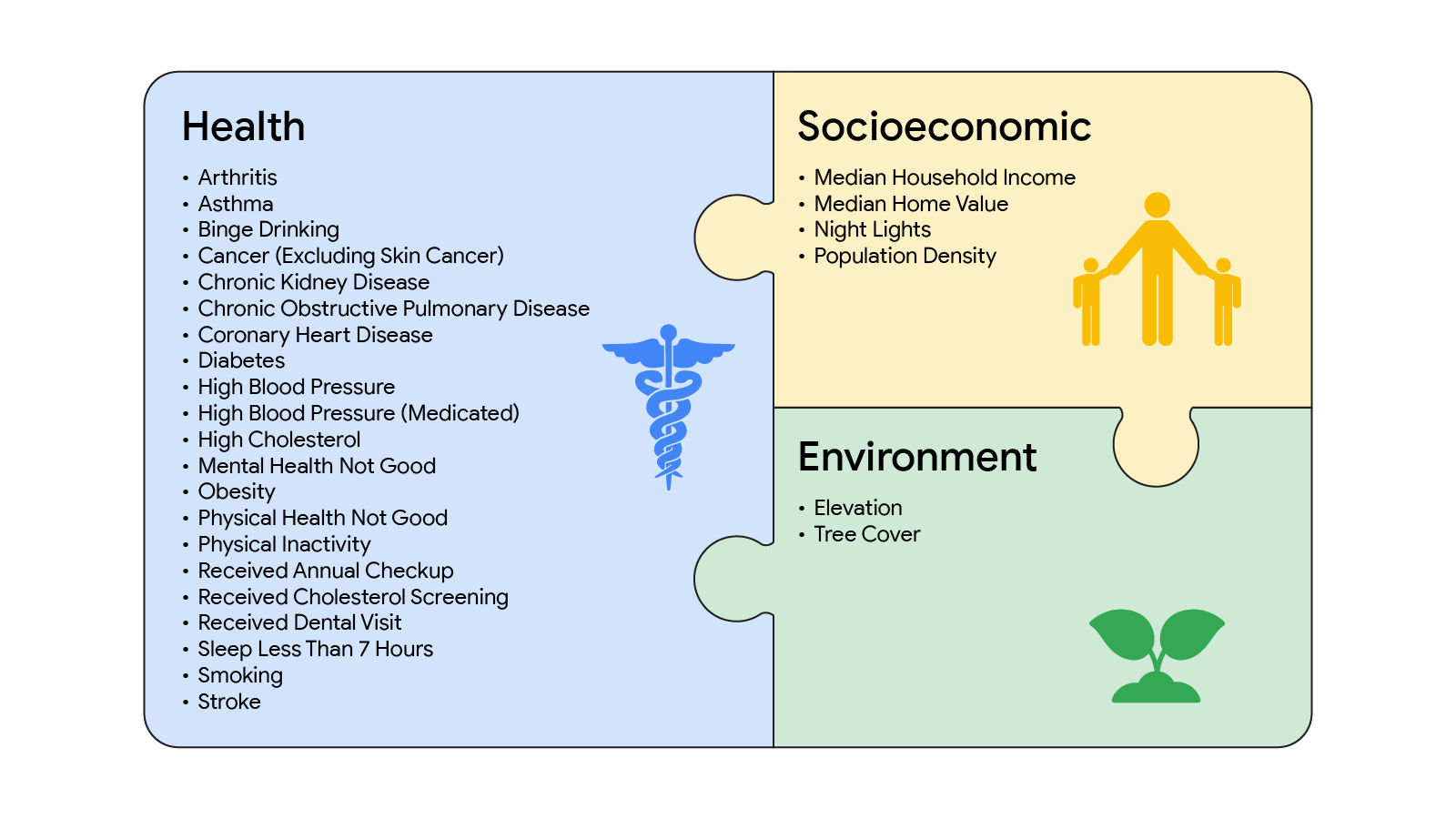}
    \caption{\footnotesize\textbf{Downstream Tasks Used for Evaluation} We show the full set of tasks we use to evaluate PDFM. This includes 21 health, 6 socioeconomic (4 used for spatial problems and 2 used for forecasting problems), and 2 environment focused tasks. Each of the 27 spatial tasks and the poverty task are available at the postal code and county level across the contiguous US. *Poverty and unemployment are only used in forecasting experiments and unemployment is only available at the county level.}
    \label{fig:categories}
\end{figure}

To develop PDFM, we gathered and organized five datasets (encompassing maps, busyness, aggregated search trends, weather and air quality) at the postal code and county levels. We then leveraged GNNs to train a foundation model that is able to generate general purpose features, or embeddings. These embeddings are used as input into simple downstream models to solve 27 downstream tasks comprising health, socioeconomic, and environmental categories. See Figure \ref{fig:categories} for the detailed list and their corresponding categories. We describe how we selected this benchmark and existing work that uses it in Section \ref{sec:benchmark_datasets} in our methods. PDFM-based embeddings were generated and used to improve an existing time series foundation model for geospatial forecasting.

To assess performance, we conducted interpolation and extrapolation experiments at the postal code level, representing real-world applications with missing data. For interpolation, we randomly chose 20\% of counties and all their corresponding postal codes as the holdout set. To simulate larger gaps in data, we performed extrapolation by selecting 20\% of US states and all their associated postal codes as the holdout set. For the super-resolution experiments, we used only county-level labels for training and 20\% of postal codes as the holdout test set.

We compared our model with inverse distance weighting (IDW) from \cite{shepard1968two} because our experiments showed this to be the most competitive method for interpolation, extrapolation, and super-resolution across techniques that only made use of geographic coordinates. We also compared performance with SatCLIP and GeoCLIP because recent work shows these location encoders being state-of-the-art on environmental and socioeconomic tasks. We include tables breaking down performance by each of the individual datasets used in the PDFM to showcase the complementary information being modeled.

\subsection{Interpolation}

\begin{figure} 
    \centering
    \includegraphics[width=\textwidth]{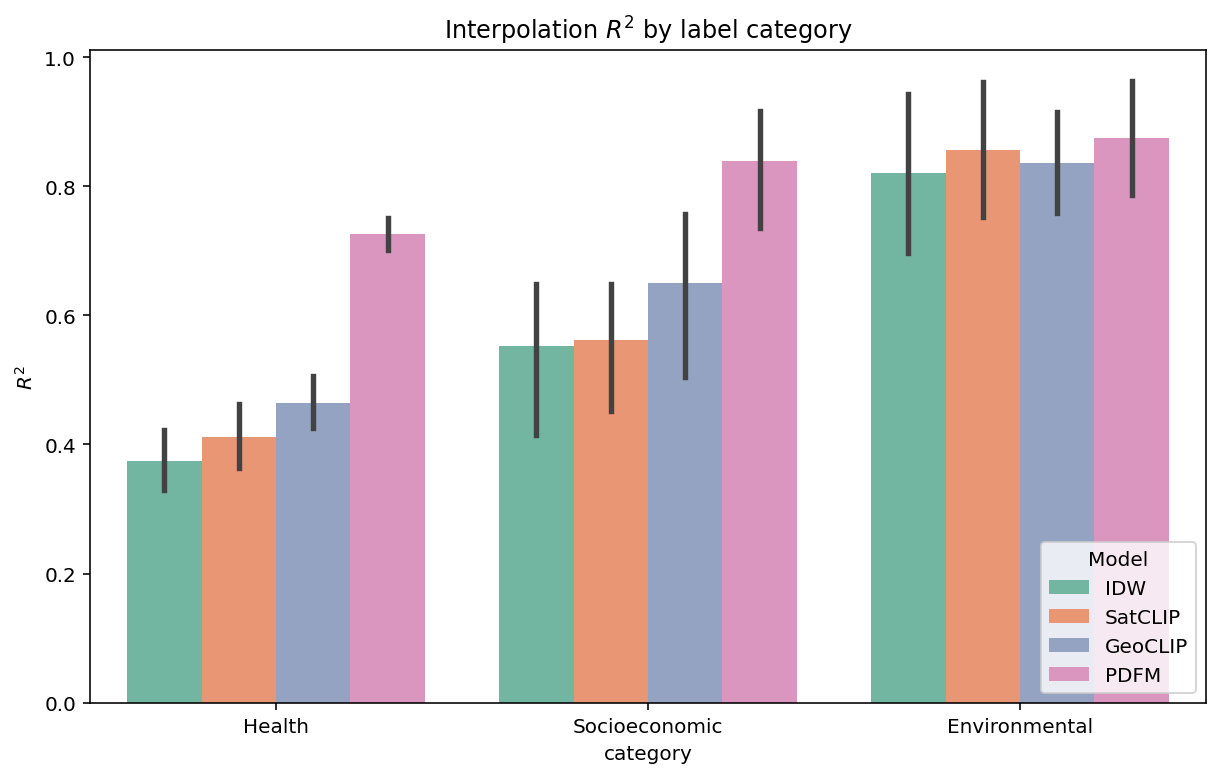}
    \caption{\footnotesize\textbf{Interpolation $R^2$ Results (Higher is Better).} We summarize the full set of experiments for interpolation across all 27 tasks using the $R^2$ metric (a higher value shows the respective model better accounts for the variance in the ground truth labels). These experiments compare the performance of inverse distance weighted (IDW) interpolation SatCLIP embeddings, GeoCLIP embeddings, and our PDFM embeddings, using GBDT as the downstream model.}
    \label{fig:bar_graph_interpolation}
\end{figure}

In Figure \ref{fig:bar_graph_interpolation}, we present our full set of experiments for interpolation across all of our 27 tasks grouped into health, socioeconomic, and environment categories using the $R^2$ metric (a higher value shows the respective model better accounts for the variance in the ground truth target variable labels). These experiments compare the performance of inverse distance weighted (IDW) interpolation SatCLIP embeddings, GeoCLIP embeddings, and our PDFM embeddings, using GBDT as the downstream model. See Appendix Table \ref{table:interpolation_full} for detailed individual task results.

Table \ref{table:interpolation} showcases the effectiveness of PDFM for interpolation across 27 health, socio-economic and environmental tasks. PDFM consistently demonstrates superior performance, achieving a mean $R^2$ of 0.83 for all 27 tasks, and 0.73 for all 21 health-related tasks. Figure \ref{fig:bar_graph_interpolation} demonstrates that PDFM significantly outperforms SatCLIP and GeoCLIP in the socioeconomic and health category. Ablation studies performed on the individual modalities highlight the strong prediction capabilities of aggregated web search trends (mean $R^2$ = 0.75), particularly for home values ($R^2$ = 0.81), night lights ($R^2$ = 0.88) and population density ($R^2$ = 0.84) measures. Maps also show a strong ability to predict nightlights ($R^2$ = 0.91) and population density ($R^2$ = 0.88). While IDW performs remarkably well in predicting elevation ($R^2$ = 0.94), both PDFM and SatCLIP achieve the highest accuracy ($R^2$ = 0.96). 

\begin{table}
\caption{\textbf{Interpolation $R^2$ Results (Higher is Better).} We summarize the full set of experiments for interpolation across all 27 tasks using the $R^2$ metric (a higher value shows the respective model better accounts for the variance in the ground truth labels). These experiments compare the performance of inverse distance weighted (IDW) interpolation SatCLIP embeddings, GeoCLIP embeddings, our PDFM embeddings and its subcomponents (Weather \& Air Quality, Aggregated Search Trends, Maps, and Busyness) using GBDT as the downstream model.}
\label{table:interpolation}
\begin{tabular}{lrrrrrrr}
 & IDW & SatCLIP & GeoCLIP & Weather \& AQ & Trends & Maps & PDFM \\
Income & {\cellcolor[HTML]{F7FCF5}} \color[HTML]{000000} 0.34 & {\cellcolor[HTML]{EFF9EB}} \color[HTML]{000000} 0.40 & {\cellcolor[HTML]{EBF7E7}} \color[HTML]{000000} 0.43 & {\cellcolor[HTML]{ECF8E8}} \color[HTML]{000000} 0.42 & {\cellcolor[HTML]{B4E1AD}} \color[HTML]{000000} 0.66 & {\cellcolor[HTML]{BAE3B3}} \color[HTML]{000000} 0.64 & \bfseries {\cellcolor[HTML]{AEDEA7}} \color[HTML]{000000} 0.68 \\
HomeValue & {\cellcolor[HTML]{F7FCF5}} \color[HTML]{000000} 0.66 & {\cellcolor[HTML]{F2FAEF}} \color[HTML]{000000} 0.68 & {\cellcolor[HTML]{F7FCF5}} \color[HTML]{000000} 0.66 & {\cellcolor[HTML]{F7FCF5}} \color[HTML]{000000} 0.66 & {\cellcolor[HTML]{BEE5B8}} \color[HTML]{000000} 0.81 & {\cellcolor[HTML]{D7EFD1}} \color[HTML]{000000} 0.76 & \bfseries {\cellcolor[HTML]{AEDEA7}} \color[HTML]{000000} 0.84 \\
NightLights & {\cellcolor[HTML]{F5FBF2}} \color[HTML]{000000} 0.57 & {\cellcolor[HTML]{F7FCF5}} \color[HTML]{000000} 0.55 & {\cellcolor[HTML]{D3EECD}} \color[HTML]{000000} 0.78 & {\cellcolor[HTML]{CEECC8}} \color[HTML]{000000} 0.80 & {\cellcolor[HTML]{BBE4B4}} \color[HTML]{000000} 0.88 & {\cellcolor[HTML]{B4E1AD}} \color[HTML]{000000} 0.91 & \bfseries {\cellcolor[HTML]{AEDEA7}} \color[HTML]{000000} 0.93 \\
PopulationDensity & {\cellcolor[HTML]{F4FBF1}} \color[HTML]{000000} 0.64 & {\cellcolor[HTML]{F7FCF5}} \color[HTML]{000000} 0.62 & {\cellcolor[HTML]{E1F3DC}} \color[HTML]{000000} 0.74 & {\cellcolor[HTML]{D6EFD0}} \color[HTML]{000000} 0.78 & {\cellcolor[HTML]{C3E7BC}} \color[HTML]{000000} 0.84 & {\cellcolor[HTML]{B5E1AE}} \color[HTML]{000000} 0.88 & \bfseries {\cellcolor[HTML]{AEDEA7}} \color[HTML]{000000} 0.90 \\
TreeCover & {\cellcolor[HTML]{DDF2D8}} \color[HTML]{000000} 0.70 & {\cellcolor[HTML]{C6E8BF}} \color[HTML]{000000} 0.75 & {\cellcolor[HTML]{C0E6B9}} \color[HTML]{000000} 0.76 & {\cellcolor[HTML]{CFECC9}} \color[HTML]{000000} 0.73 & {\cellcolor[HTML]{F7FCF5}} \color[HTML]{000000} 0.62 & {\cellcolor[HTML]{EFF9EB}} \color[HTML]{000000} 0.65 & \bfseries {\cellcolor[HTML]{AEDEA7}} \color[HTML]{000000} 0.79 \\
Elevation & {\cellcolor[HTML]{B6E2AF}} \color[HTML]{000000} 0.94 & \bfseries {\cellcolor[HTML]{AEDEA7}} \color[HTML]{000000} 0.96 & {\cellcolor[HTML]{C2E7BB}} \color[HTML]{000000} 0.91 & \bfseries {\cellcolor[HTML]{AEDEA7}} \color[HTML]{000000} 0.96 & {\cellcolor[HTML]{EDF8EA}} \color[HTML]{000000} 0.76 & {\cellcolor[HTML]{F7FCF5}} \color[HTML]{000000} 0.71 & \bfseries {\cellcolor[HTML]{AEDEA7}} \color[HTML]{000000} 0.96 \\
Health (mean) & {\cellcolor[HTML]{F7FCF5}} \color[HTML]{000000} 0.37 & {\cellcolor[HTML]{F2FAEF}} \color[HTML]{000000} 0.41 & {\cellcolor[HTML]{EBF7E7}} \color[HTML]{000000} 0.46 & {\cellcolor[HTML]{EBF7E7}} \color[HTML]{000000} 0.46 & {\cellcolor[HTML]{C4E8BD}} \color[HTML]{000000} 0.65 & {\cellcolor[HTML]{C7E9C0}} \color[HTML]{000000} 0.64 & \bfseries {\cellcolor[HTML]{AEDEA7}} \color[HTML]{000000} 0.73 \\
All metrics (mean) & {\cellcolor[HTML]{F7FCF5}} \color[HTML]{000000} 0.60 & {\cellcolor[HTML]{F3FAF0}} \color[HTML]{000000} 0.62 & {\cellcolor[HTML]{E7F6E2}} \color[HTML]{000000} 0.68 & {\cellcolor[HTML]{E4F5DF}} \color[HTML]{000000} 0.69 & {\cellcolor[HTML]{D3EECD}} \color[HTML]{000000} 0.74 & {\cellcolor[HTML]{D3EECD}} \color[HTML]{000000} 0.74 & \bfseries {\cellcolor[HTML]{AEDEA7}} \color[HTML]{000000} 0.83 \\
\end{tabular}
\end{table}

\begin{figure} 
    \centering
    \includegraphics[width=\textwidth]{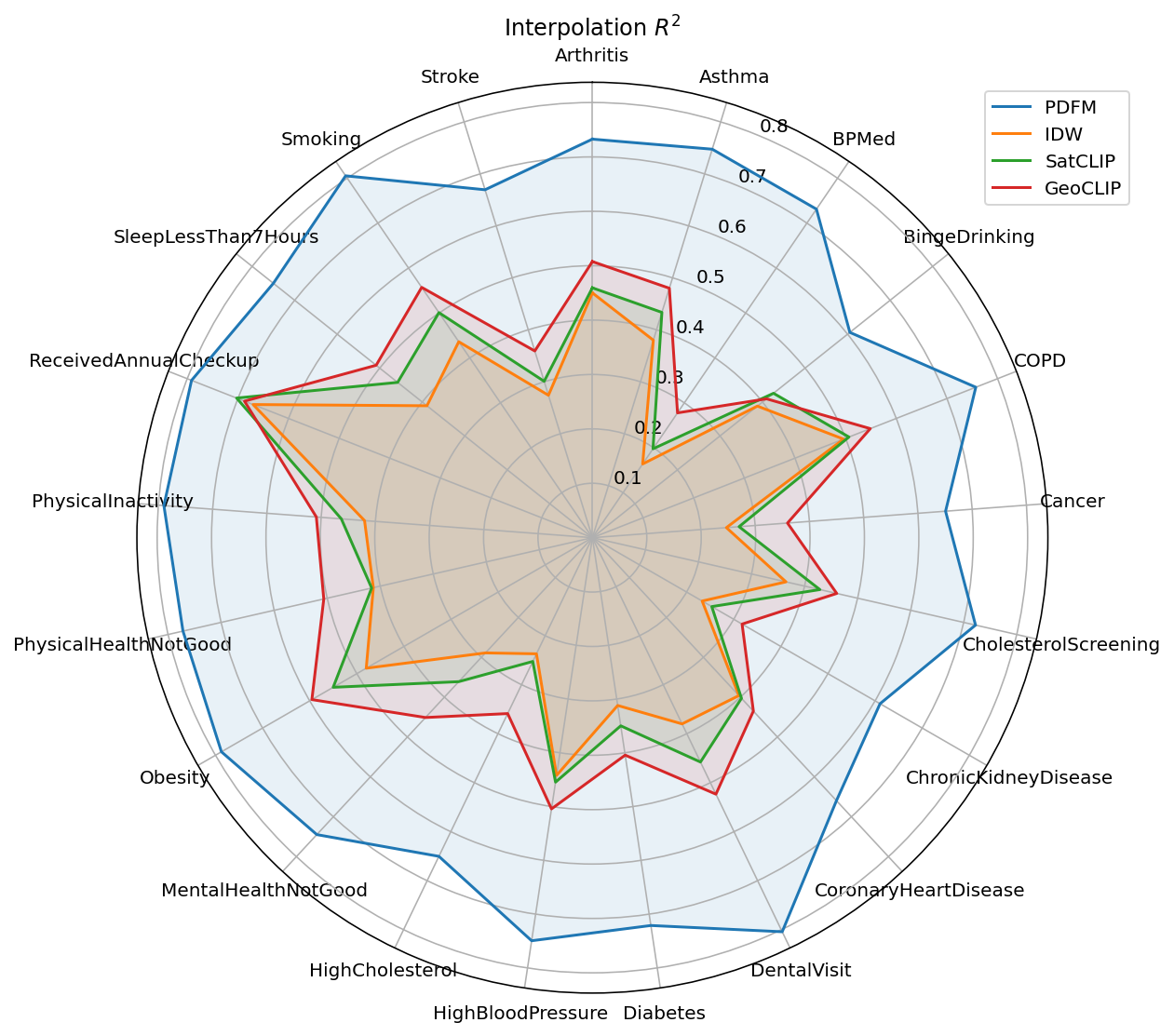}
    \caption{\footnotesize\textbf{Interpolation Performance Across Methods for Health Tasks} This radar plot shows interpolation R2 performance of PDFM across all 21 health related tasks compared with IDW, SatCLIP, and GeoCLIP.}
    \label{fig:radar_interpolation}
\end{figure}

\begin{figure} 
    \centering
    \includegraphics[width=\textwidth]{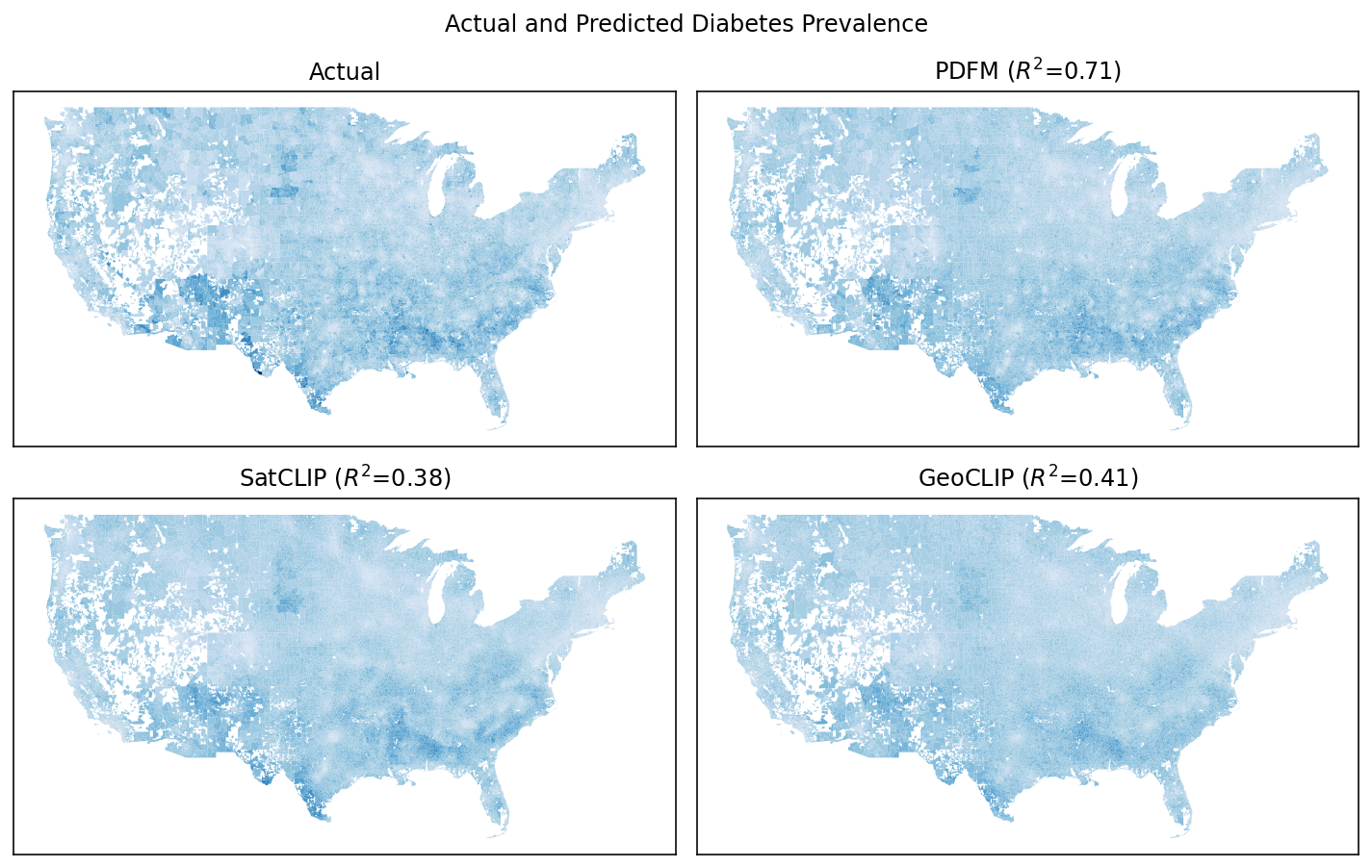}
    \caption{\footnotesize\textbf{Diabetes Prevalence Interpolation Visualized Across Methods} Actual and predicted Diabetes prevalence in CONUS.}
    \label{fig:us_diabetes}
\end{figure}

\begin{figure}
    \centering
    \includegraphics[width=\textwidth]{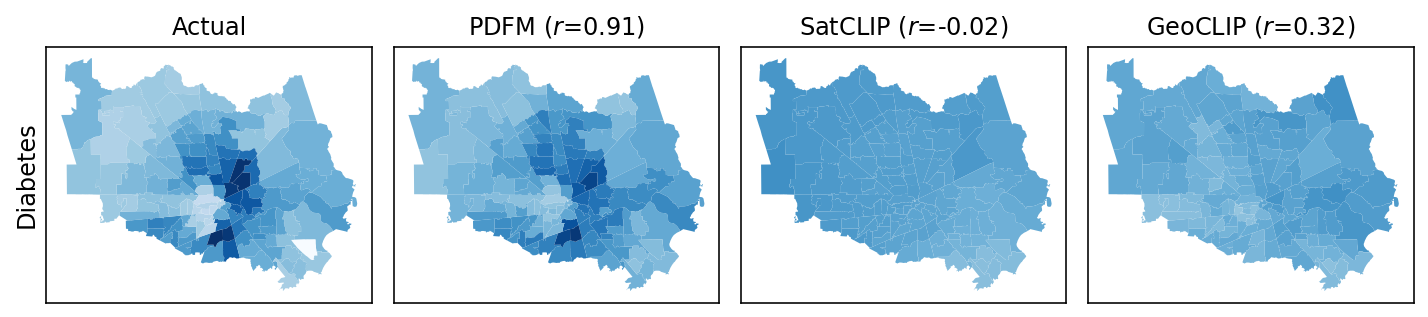}
    \includegraphics[width=\textwidth]{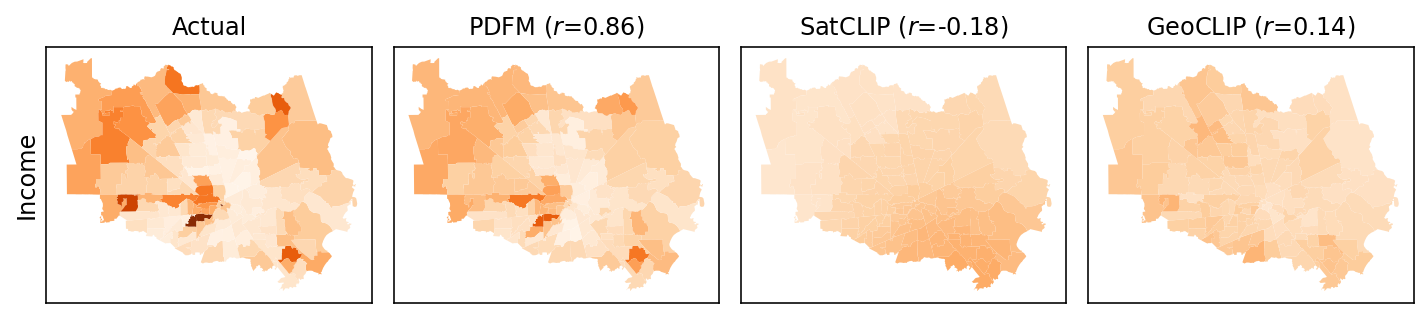}
    \includegraphics[width=\textwidth]{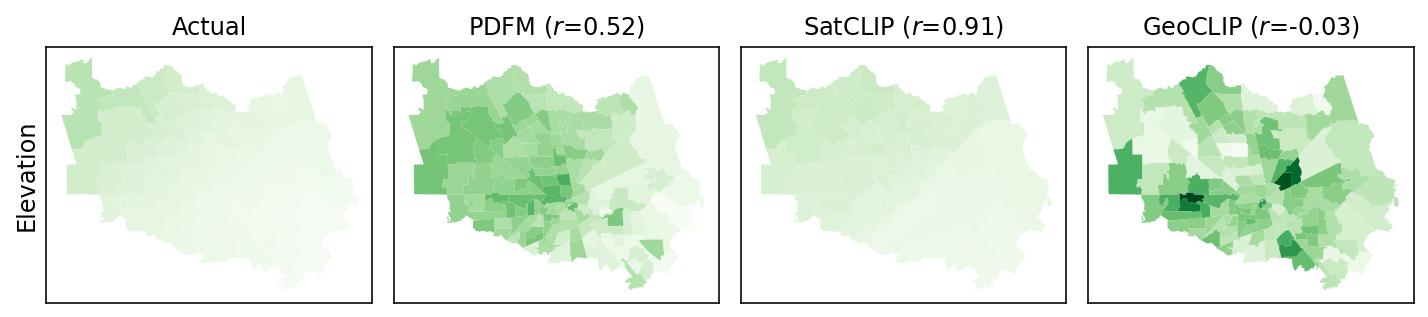}
    \caption{\footnotesize\textbf{Comparison of Postal Code Level Predictions} Actual and predicted Diabetes, Income, and Elevation values in the 131 postal codes in Harris County, TX from the holdout set.}
    \label{fig:county_visual}
\end{figure}

Figure \ref{fig:us_diabetes} shows the actual and predicted diabetes prevalence in all 32K postal codes in CONUS. All predicted values are from 5-fold cross validation and test, no training set predictions are depicted. While all three models are able to capture broad national trends, PDFM predictions capture more diverse details over small regions. This distinction is more clearly illustrated when we zoom in and compare the predictions for postal codes in a single holdout set county in Figure \ref{fig:county_visual}. We report Pearson's correlation coefficient $r$ for intra-county evaluation because $R^2$ is not well defined for subset evaluation.

It is notable that SatCLIP features perform remarkably well for predicting Elevation, which tend to vary gradually over larger regions and does not require high spatial resolution.

\subsection{Extrapolation}

\begin{figure}
    \centering
    \includegraphics[width=\textwidth]{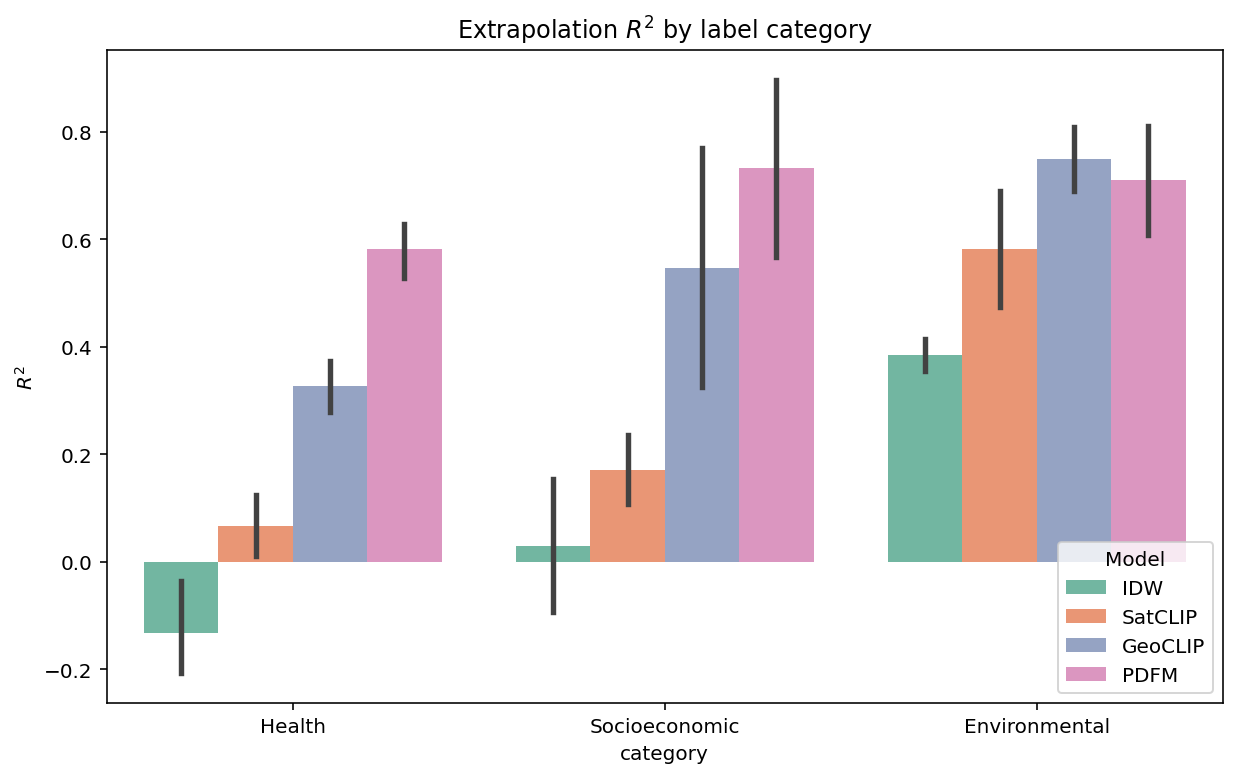}
    \caption{\footnotesize\textbf{Extrapolation $R^2$ Results (Higher is Better).} We present our full set of experiments for extrapolation across all 27 tasks grouped into health, socioeconomic, and environment categories using the R2 metric (a higher value shows the respective model better accounts for the variance in the ground truth labels). These experiments compare the performance of inverse distance weighted (IDW) interpolation SatCLIP embeddings, GeoCLIP embeddings and our PDFM embeddings using GBDT as the downstream model.}
    \label{fig:bar_graph_extrapolation}
\end{figure}

In Figure \ref{fig:bar_graph_extrapolation}, we present our full set of experiments for the extrapolation task across all 27 downstream tasks grouped into health, socioeconomic, and environment categories using the $R^2$ metric (a higher value shows the respective model better accounts for the variance in the ground truth labels). These experiments compare the performance of inverse distance weighted (IDW) interpolation, SatCLIP embeddings, GeoCLIP embeddings, and our PDFM embeddings using GBDT as the downstream model. See Table \ref{table:extrapolation_full} for detailed individual target label results across all experiment variants for the extrapolation task.

\begin{table}
\caption{\textbf{Extrapolation $R^2$ Results (Higher is Better).} We present our full set of experiments for extrapolation across all 27 tasks using the $R^2$ metric (a higher value shows the respective model better accounts for the variance in the ground truth labels). These experiments compare the performance of inverse distance weighted (IDW) interpolation, SatCLIP embeddings, GeoCLIP embeddings, our PDFM embeddings and its subcomponents (Weather \& Air Quality, Aggregated Search Trends, Maps, and Busyness) using GBDT as the downstream model.}
\label{table:extrapolation}
\begin{tabular}{lrrrrrrr}
 & IDW & SatCLIP & GeoCLIP & Weather \& AQ & Trends & Maps & PDFM \\
Income & {\cellcolor[HTML]{F7FCF5}} \color[HTML]{000000} -0.09 & {\cellcolor[HTML]{E9F7E5}} \color[HTML]{000000} 0.11 & {\cellcolor[HTML]{CFECC9}} \color[HTML]{000000} 0.35 & {\cellcolor[HTML]{E5F5E0}} \color[HTML]{000000} 0.17 & \bfseries {\cellcolor[HTML]{AEDEA7}} \color[HTML]{000000} 0.59 & {\cellcolor[HTML]{B1E0AB}} \color[HTML]{000000} 0.57 & \bfseries {\cellcolor[HTML]{AEDEA7}} \color[HTML]{000000} 0.59 \\
HomeValue & {\cellcolor[HTML]{F7FCF5}} \color[HTML]{000000} -0.09 & {\cellcolor[HTML]{E9F7E5}} \color[HTML]{000000} 0.10 & {\cellcolor[HTML]{D3EECD}} \color[HTML]{000000} 0.29 & {\cellcolor[HTML]{E0F3DB}} \color[HTML]{000000} 0.19 & {\cellcolor[HTML]{B1E0AB}} \color[HTML]{000000} 0.52 & {\cellcolor[HTML]{B1E0AB}} \color[HTML]{000000} 0.52 & \bfseries {\cellcolor[HTML]{AEDEA7}} \color[HTML]{000000} 0.54 \\
NightLights & {\cellcolor[HTML]{F7FCF5}} \color[HTML]{000000} 0.11 & {\cellcolor[HTML]{EFF9EC}} \color[HTML]{000000} 0.25 & {\cellcolor[HTML]{BCE4B5}} \color[HTML]{000000} 0.80 & {\cellcolor[HTML]{CBEAC4}} \color[HTML]{000000} 0.68 & {\cellcolor[HTML]{B6E2AF}} \color[HTML]{000000} 0.85 & {\cellcolor[HTML]{B0DFAA}} \color[HTML]{000000} 0.89 & \bfseries {\cellcolor[HTML]{AEDEA7}} \color[HTML]{000000} 0.91 \\
PopulationDensity & {\cellcolor[HTML]{F7FCF5}} \color[HTML]{000000} 0.20 & {\cellcolor[HTML]{F6FCF4}} \color[HTML]{000000} 0.22 & {\cellcolor[HTML]{C3E7BC}} \color[HTML]{000000} 0.74 & {\cellcolor[HTML]{CDECC7}} \color[HTML]{000000} 0.66 & {\cellcolor[HTML]{BBE4B4}} \color[HTML]{000000} 0.79 & {\cellcolor[HTML]{AFDFA8}} \color[HTML]{000000} 0.87 & \bfseries {\cellcolor[HTML]{AEDEA7}} \color[HTML]{000000} 0.88 \\
TreeCover & {\cellcolor[HTML]{F7FCF5}} \color[HTML]{000000} 0.36 & {\cellcolor[HTML]{E7F6E3}} \color[HTML]{000000} 0.47 & \bfseries {\cellcolor[HTML]{AEDEA7}} \color[HTML]{000000} 0.69 & {\cellcolor[HTML]{DCF2D7}} \color[HTML]{000000} 0.52 & {\cellcolor[HTML]{EAF7E6}} \color[HTML]{000000} 0.45 & {\cellcolor[HTML]{DCF2D7}} \color[HTML]{000000} 0.52 & {\cellcolor[HTML]{C7E9C0}} \color[HTML]{000000} 0.61 \\
Elevation & {\cellcolor[HTML]{F7FCF5}} \color[HTML]{000000} 0.41 & {\cellcolor[HTML]{CBEBC5}} \color[HTML]{000000} 0.69 & \bfseries {\cellcolor[HTML]{AEDEA7}} \color[HTML]{000000} 0.81 & {\cellcolor[HTML]{B2E0AC}} \color[HTML]{000000} 0.79 & {\cellcolor[HTML]{E9F7E5}} \color[HTML]{000000} 0.53 & {\cellcolor[HTML]{EAF7E6}} \color[HTML]{000000} 0.52 & \bfseries {\cellcolor[HTML]{AEDEA7}} \color[HTML]{000000} 0.81 \\
Health (mean) & {\cellcolor[HTML]{F7FCF5}} \color[HTML]{000000} -0.13 & {\cellcolor[HTML]{E9F7E5}} \color[HTML]{000000} 0.07 & {\cellcolor[HTML]{CFECC9}} \color[HTML]{000000} 0.33 & {\cellcolor[HTML]{DEF2D9}} \color[HTML]{000000} 0.20 & {\cellcolor[HTML]{BBE4B4}} \color[HTML]{000000} 0.49 & {\cellcolor[HTML]{B6E2AF}} \color[HTML]{000000} 0.52 & \bfseries {\cellcolor[HTML]{AEDEA7}} \color[HTML]{000000} 0.58 \\
All metrics (mean) & {\cellcolor[HTML]{F7FCF5}} \color[HTML]{000000} 0.11 & {\cellcolor[HTML]{EAF7E6}} \color[HTML]{000000} 0.27 & {\cellcolor[HTML]{C4E8BD}} \color[HTML]{000000} 0.57 & {\cellcolor[HTML]{D4EECE}} \color[HTML]{000000} 0.46 & {\cellcolor[HTML]{C0E6B9}} \color[HTML]{000000} 0.60 & {\cellcolor[HTML]{BAE3B3}} \color[HTML]{000000} 0.63 & \bfseries {\cellcolor[HTML]{AEDEA7}} \color[HTML]{000000} 0.70 \\
\end{tabular}
\end{table}

Extrapolation remains a challenging task due to significant gaps in labeled data. In this context, PDFM demonstrates (Table \ref{table:extrapolation}) superior performance with an $R^2$ value of 0.70 across all metrics and 0.58 for health-related metrics. Leveraging geotagged images, GeoCLIP excels in identifying tree cover, achieving an $R^2$ value of 0.69, surpassing both PDFM and any single modality. Overall, PDFM outperforms benchmarks in 25 out of 27 tasks, highlighting its effectiveness in extrapolation scenarios. Figure \ref{fig:bar_graph_extrapolation} illustrates that while GeoCLIP shows a slight advantage in handling environmental tasks, PDFM significantly outperforms all other baselines when it comes to predicting health and socio-economic variables.

\begin{figure}
    \centering
    \includegraphics[width=\textwidth]{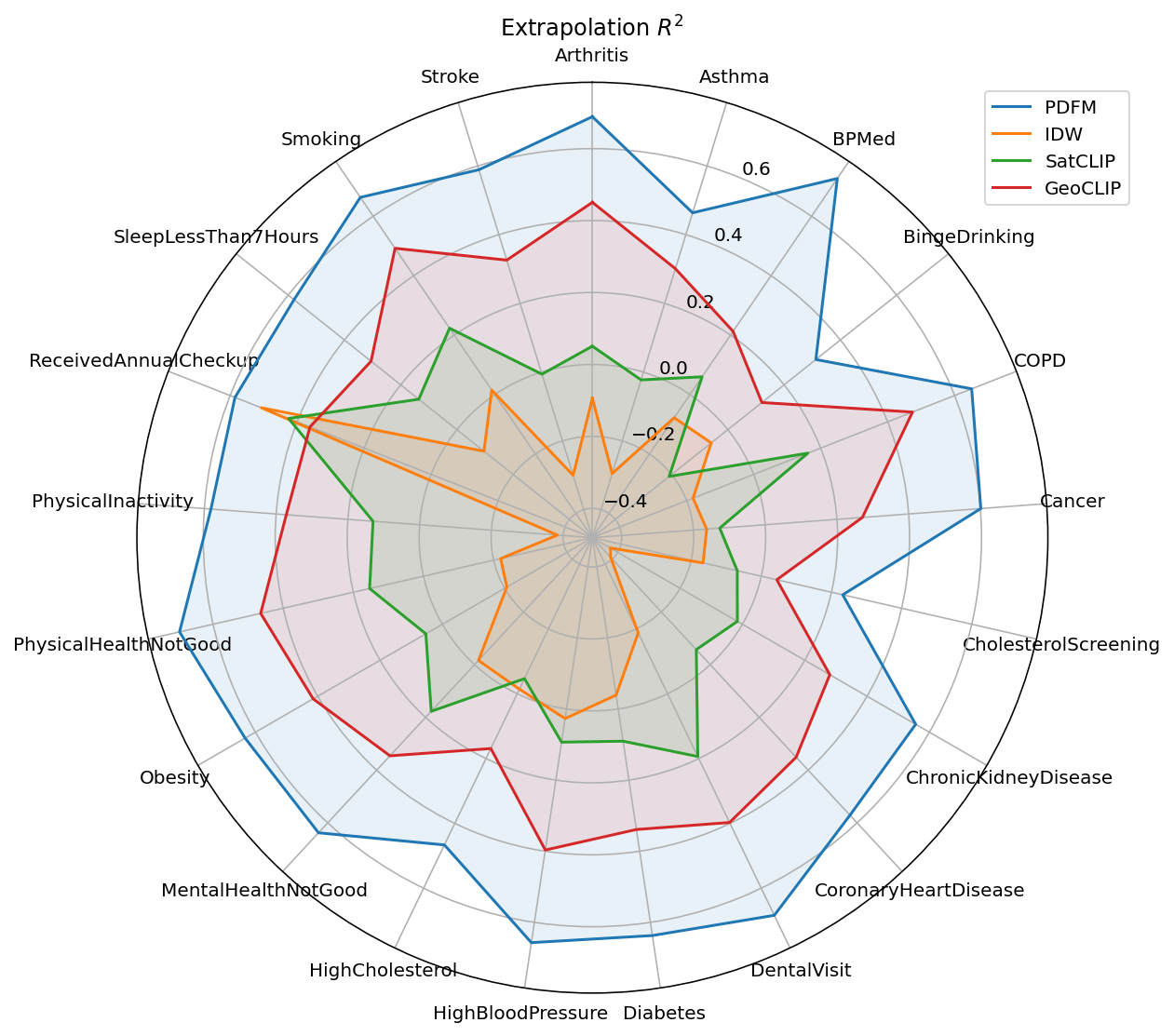}
    \caption{\footnotesize\textbf{Extrapolation Performance Across Methods for Health Tasks.} This radar plot shows extrapolation $R^2$ performance of PDFM across all 21 health related tasks compared with IDW, SatCLIP, and GeoCLIP.}
    \label{fig:radar_extrapolation}
\end{figure}

\subsection{Super-Resolution}

\begin{figure}
    \centering
    \includegraphics[width=\textwidth]{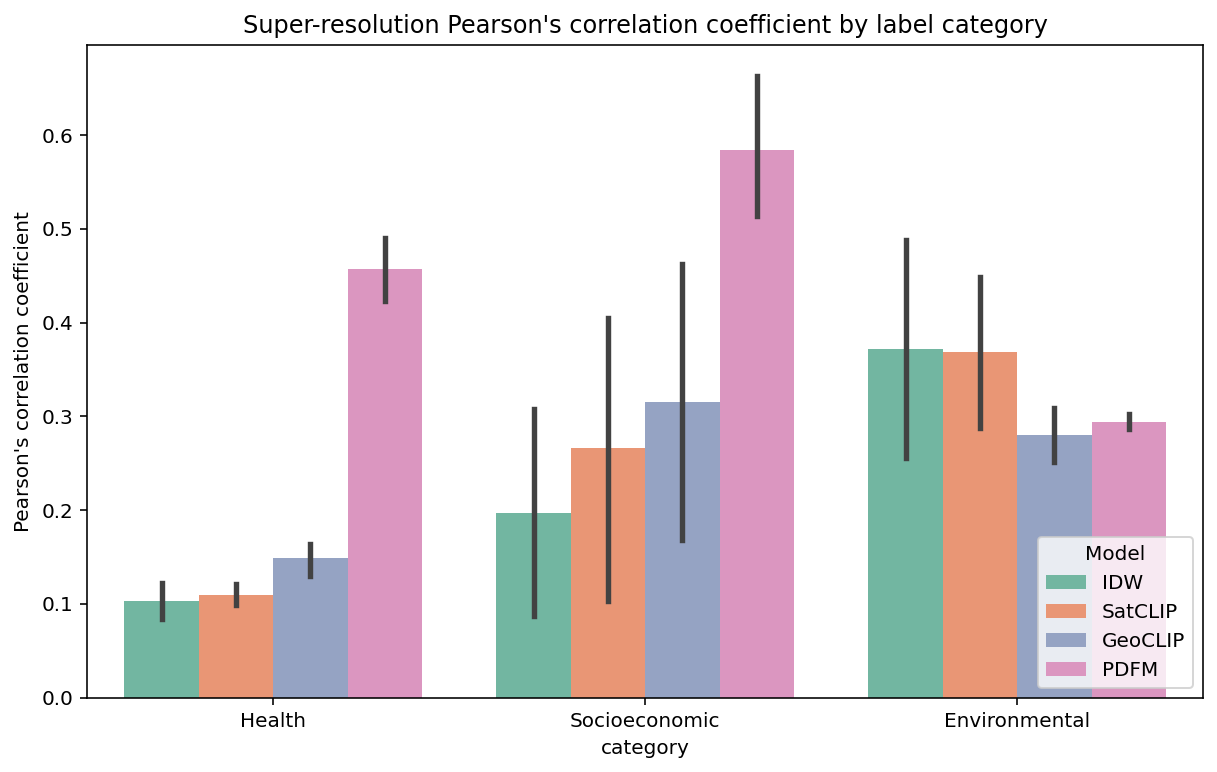}
    \caption{\footnotesize\textbf{Intracounty Super-Resolution Pearson's Correlation Coefficients ($r$)  (Higher is Better).} We present the full set of experiments for the super-resolution task across all 27 labels using the Pearson's $r$ metric (a higher value means the respective model's predictions are better correlated with the ground truth postal code labels in each county). These experiments compare the performance of inverse distance weighted (IDW) interpolation SatCLIP embeddings, GeoCLIP embeddings, and our PDFM embeddings.}
    \label{fig:bar_graph_superresolution}
\end{figure}

In Figure \ref{fig:bar_graph_superresolution}, we present the full set of experiments for the super-resolution task across all of 27 labels grouped into health, socioeconomic, and environment categories using the mean intra-county Pearson's correlation coefficient (a higher value means the respective model's predictions are better correlated with the ground truth postal code-level labels in each county). These experiments compare the performance of inverse distance weighted (IDW) interpolation, SatCLIP embeddings, GeoCLIP embeddings, and our PDFM embeddings. See Appendix Table \ref{table:superresolution_full} for detailed individual target label results across all experiment variants for the super-resolution task.

\begin{table}
\caption{\textbf{Intracounty Super-Resolution Pearson's Correlation Coefficients (r) (Higher is Better).} We present the full set of experiments for the super-resolution task across all 27 labels using the Pearson's r metric (a higher value means the respective model's predictions are better correlated with the ground truth postal code labels in each county). These experiments compare the performance of inverse distance weighted (IDW) interpolation SatCLIP embeddings, GeoCLIP embeddings, our PDFM embeddings and its subcomponents (Weather \& Air Quality, Aggregated Search Trends, Maps, and Busyness).}
\label{table:superresolution}
\begin{tabular}{lrrrrrrr}
 & IDW & SatCLIP & GeoCLIP & Weather \& AQ & Trends & Maps & PDFM \\
Income & {\cellcolor[HTML]{F3FAF0}} \color[HTML]{000000} 0.06 & {\cellcolor[HTML]{F1FAEE}} \color[HTML]{000000} 0.08 & {\cellcolor[HTML]{E9F7E5}} \color[HTML]{000000} 0.16 & {\cellcolor[HTML]{F7FCF5}} \color[HTML]{000000} 0.02 & {\cellcolor[HTML]{B0DFAA}} \color[HTML]{000000} 0.49 & {\cellcolor[HTML]{B8E3B2}} \color[HTML]{000000} 0.45 & \bfseries {\cellcolor[HTML]{AEDEA7}} \color[HTML]{000000} 0.50 \\
HomeValue & {\cellcolor[HTML]{EFF9EB}} \color[HTML]{000000} 0.11 & {\cellcolor[HTML]{E8F6E3}} \color[HTML]{000000} 0.18 & {\cellcolor[HTML]{E8F6E3}} \color[HTML]{000000} 0.18 & {\cellcolor[HTML]{F7FCF5}} \color[HTML]{000000} 0.02 & \bfseries {\cellcolor[HTML]{AEDEA7}} \color[HTML]{000000} 0.52 & {\cellcolor[HTML]{BCE4B5}} \color[HTML]{000000} 0.45 & \bfseries {\cellcolor[HTML]{AEDEA7}} \color[HTML]{000000} 0.52 \\
NightLights & {\cellcolor[HTML]{F7FCF5}} \color[HTML]{000000} 0.28 & {\cellcolor[HTML]{EAF7E6}} \color[HTML]{000000} 0.40 & {\cellcolor[HTML]{DCF2D7}} \color[HTML]{000000} 0.49 & {\cellcolor[HTML]{EBF7E7}} \color[HTML]{000000} 0.39 & {\cellcolor[HTML]{BAE3B3}} \color[HTML]{000000} 0.66 & {\cellcolor[HTML]{C3E7BC}} \color[HTML]{000000} 0.62 & \bfseries {\cellcolor[HTML]{AEDEA7}} \color[HTML]{000000} 0.71 \\
PopulationDensity & {\cellcolor[HTML]{F7FCF5}} \color[HTML]{000000} 0.33 & {\cellcolor[HTML]{E9F7E5}} \color[HTML]{000000} 0.41 & {\cellcolor[HTML]{E5F5E1}} \color[HTML]{000000} 0.43 & {\cellcolor[HTML]{F0F9ED}} \color[HTML]{000000} 0.37 & {\cellcolor[HTML]{B5E1AE}} \color[HTML]{000000} 0.58 & {\cellcolor[HTML]{DAF0D4}} \color[HTML]{000000} 0.47 & \bfseries {\cellcolor[HTML]{AEDEA7}} \color[HTML]{000000} 0.60 \\
TreeCover & {\cellcolor[HTML]{CCEBC6}} \color[HTML]{000000} 0.26 & {\cellcolor[HTML]{BBE4B4}} \color[HTML]{000000} 0.29 & \bfseries {\cellcolor[HTML]{AEDEA7}} \color[HTML]{000000} 0.31 & {\cellcolor[HTML]{E8F6E4}} \color[HTML]{000000} 0.20 & {\cellcolor[HTML]{F7FCF5}} \color[HTML]{000000} 0.15 & {\cellcolor[HTML]{E0F3DB}} \color[HTML]{000000} 0.22 & {\cellcolor[HTML]{B4E1AD}} \color[HTML]{000000} 0.30 \\
Elevation & \bfseries {\cellcolor[HTML]{AEDEA7}} \color[HTML]{000000} 0.49 & {\cellcolor[HTML]{B6E2AF}} \color[HTML]{000000} 0.45 & {\cellcolor[HTML]{DCF2D7}} \color[HTML]{000000} 0.25 & {\cellcolor[HTML]{D0EDCA}} \color[HTML]{000000} 0.32 & {\cellcolor[HTML]{F4FBF2}} \color[HTML]{000000} 0.05 & {\cellcolor[HTML]{F7FCF5}} \color[HTML]{000000} 0.02 & {\cellcolor[HTML]{D5EFCF}} \color[HTML]{000000} 0.29 \\
Health (mean) & {\cellcolor[HTML]{F5FBF2}} \color[HTML]{000000} 0.10 & {\cellcolor[HTML]{F4FBF1}} \color[HTML]{000000} 0.11 & {\cellcolor[HTML]{EFF9EB}} \color[HTML]{000000} 0.15 & {\cellcolor[HTML]{F7FCF5}} \color[HTML]{000000} 0.08 & {\cellcolor[HTML]{B0DFAA}} \color[HTML]{000000} 0.45 & {\cellcolor[HTML]{CEECC8}} \color[HTML]{000000} 0.33 & \bfseries {\cellcolor[HTML]{AEDEA7}} \color[HTML]{000000} 0.46 \\
All metrics (mean) & {\cellcolor[HTML]{F2FAEF}} \color[HTML]{000000} 0.23 & {\cellcolor[HTML]{EBF7E7}} \color[HTML]{000000} 0.27 & {\cellcolor[HTML]{E9F7E5}} \color[HTML]{000000} 0.28 & {\cellcolor[HTML]{F7FCF5}} \color[HTML]{000000} 0.20 & {\cellcolor[HTML]{C3E7BC}} \color[HTML]{000000} 0.42 & {\cellcolor[HTML]{D3EECD}} \color[HTML]{000000} 0.37 & \bfseries {\cellcolor[HTML]{AEDEA7}} \color[HTML]{000000} 0.48 \\
\end{tabular}
\end{table}

Super-resolution is a difficult task, where the models are trained on only county-level labels to infer postal code labels. The goal of this task is to capture the label variances of postal codes within each county rather than simply learning to distinguish between postal codes across counties, therefore we evaluate the postal codes within each county separately then compute the average metric across counties. Pearson's correlation coefficient r is used as the eval metric, and a summary of the results are presented in Table \ref{table:superresolution}. IDW performed best on the Elevation task, and GeoCLIP performed the best on Tree Cover. Overall, PDFM demonstrates superior performance across 25 out of 27 tasks, achieving an average Pearson's r metric of 0.48.

\begin{figure}
    \centering
    \includegraphics[width=\textwidth]{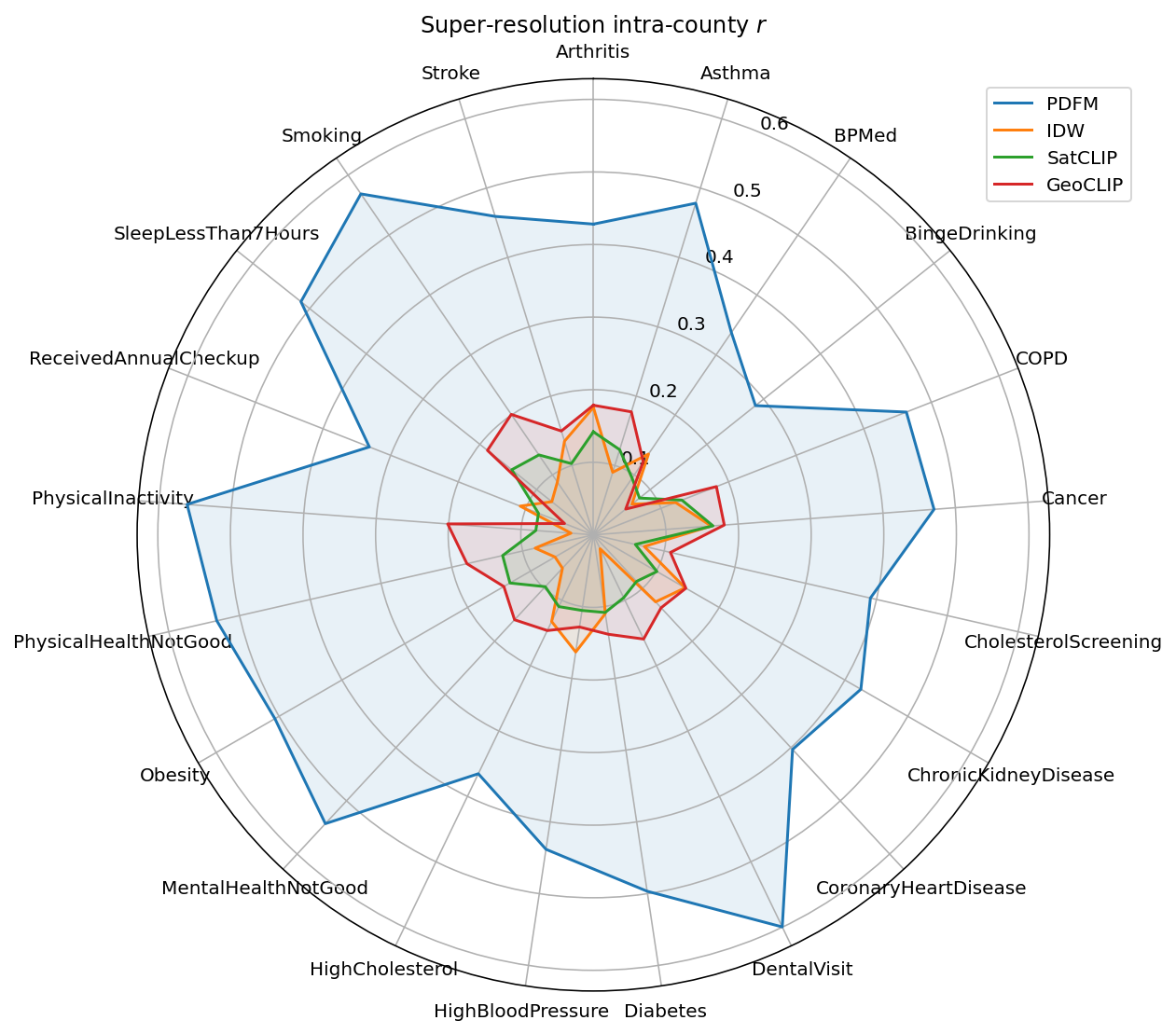}
    \caption{\footnotesize\textbf{Super-resolution Performance Across Methods for Health Tasks.} his radar plot shows super-resolution correlation performance of the PDFM across all 21 health related tasks compared with IDW, SatCLIP, and GeoCLIP.}
    \label{fig:radar_super_resolution}
\end{figure}

In Figure \ref{fig:radar_super_resolution}, we show a radar plot of super-resolution correlation performance of the PDFM across all 21 health related tasks compared with IDW, SatCLIP, and GeoCLIP.

\subsection{Model Variations}

\begin{figure} 
    \centering
    \includegraphics[width=0.8\textwidth]{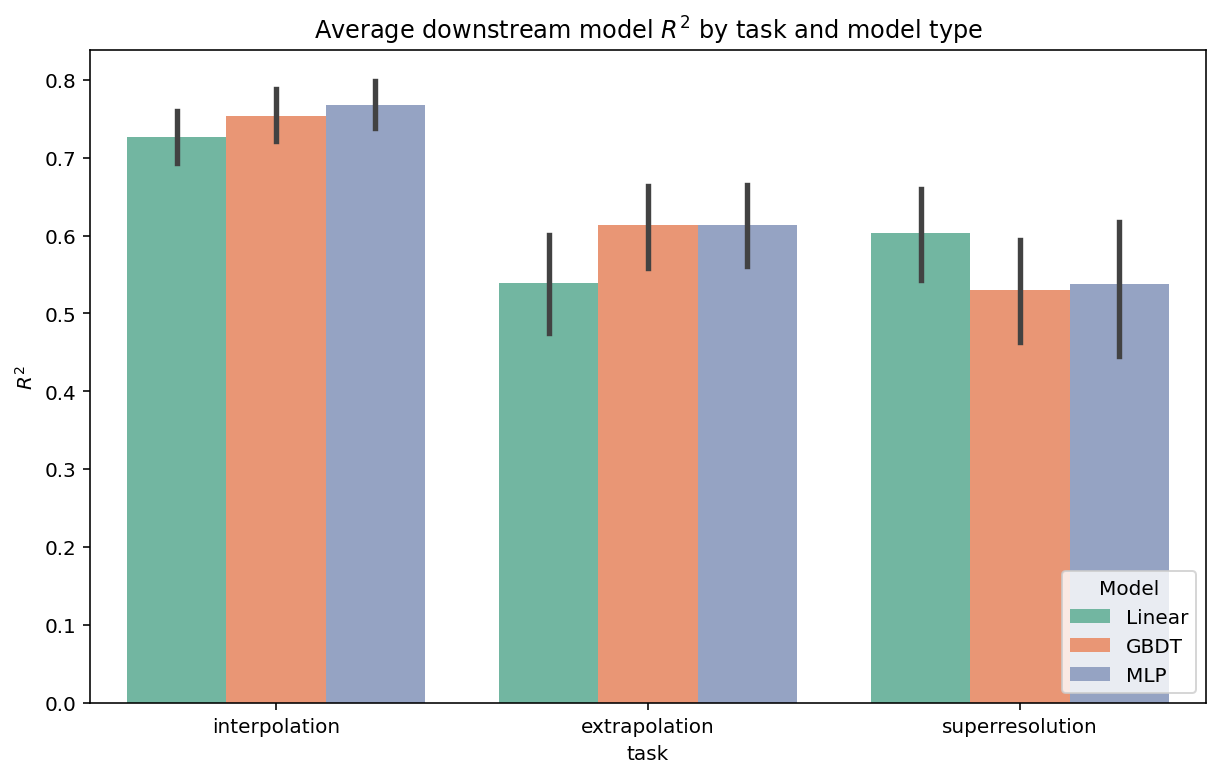}
    \caption{\footnotesize\textbf{Comparison of Downstream Model Performance} Downstream model performance using PDFM embeddings.}
    \label{fig:phase2_model_comparison}
\end{figure}

Figure \ref{fig:phase2_model_comparison} presents a comparison of PDFM embeddings' performance when used with different types of downstream models, averaged over all 27 tasks. The models included in this comparison are linear regression, gradient boosted decision trees, and multilayer perceptrons.  The results show that using a simple linear regression model with PDFM embeddings works fairly well, and more complex models improve performance except for in the case of super-resolution. For super-resolution, there are 10x less data points for training than for the other tasks, which makes it more difficult to avoid overfitting with more complex models. 

\begin{table}[h]
\caption{\textbf{Concatenating PDFM and SatCLIP Embeddings.} Adding SatCLIP embeddings as another signal to the PDFM yielded moderate performance improvements for the Interpolation and Super-resolution tasks. We present the $R^2$ metric for all tasks, where the values are in bold for the better performing model.}
\label{table:with_satclip}
\centering
\begin{tabular}{|l|c|c|c|c|c|c|}
\hline
 & \multicolumn{2}{c|}{\textbf{Interpolation}} & \multicolumn{2}{c|}{\textbf{Extrapolation}} & \multicolumn{2}{c|}{\textbf{Super-resolution}} \\
\hline
 & \textbf{PDFM} & \textbf{\begin{tabular}[c]{@{}c@{}}PDFM w/\\ SatCLIP\end{tabular}} & \textbf{PDFM} & \textbf{\begin{tabular}[c]{@{}c@{}}PDFM w/\\ SatCLIP\end{tabular}} & \textbf{PDFM} & \textbf{\begin{tabular}[c]{@{}c@{}}PDFM w/\\ SatCLIP\end{tabular}} \\
\hline
Income & 0.68 & \textbf{0.69} & 0.59 & 0.59 & 0.62 & 0.62 \\
\hline
HomeValue & 0.84 & \textbf{0.85} & \textbf{0.54} & 0.45 & 0.82 & 0.82 \\
\hline
NightLights & 0.93 & 0.93 & 0.91 & 0.91 & 0.87 & 0.87 \\
\hline
PopulationDensity & 0.90 & 0.90 & 0.88 & 0.88 & 0.80 & \textbf{0.81} \\
\hline
TreeCover & 0.79 & \textbf{0.83} & 0.61 & \textbf{0.66} & 0.64 & \textbf{0.71} \\
\hline
Elevation & 0.96 & \textbf{0.98} & \textbf{0.81} & 0.74 & 0.90 & \textbf{0.94} \\
\hline
Health (mean) & 0.73 & \textbf{0.74} & 0.58 & 0.58 & 0.55 & \textbf{0.58} \\
\hline
All metrics (mean) & 0.83 & \textbf{0.84} & \textbf{0.70} & 0.69 & 0.74 & \textbf{0.76} \\
\hline
\end{tabular}
\end{table}

In Table \ref{table:with_satclip} we compare performance across interpolation, extrapolation, and super-resolution with a version that includes the existing datasets (denoted PDFM) and a version that adds in SatCLIP features (denoted PDFM w/ SatCLIP). We found significant gains for most metrics for interpolation and more than half for super-resolution, showing that the PDFM is able to harmonize the datasets to exceed the performance of each individual modality. For extrapolation, the results are mixed, with degraded performance on some variables when the additional context from the satellite-based SatCLIP features is provided. 

\subsection{Forecasting}

\begin{table}[h]
\caption{\textbf{Forecasting.} We evaluated the benefit of using PDFM embeddings to correct the forecasting errors from TimesFM, a general purpose univariate forecasting foundation model. We train a simple adapter, a two layer multi-layer perceptron (multilayer perceptron (MLP)) to augment TimesFM forecasts using the embedding from the GNN and compare the result with a fully supervised approach (ARIMA). The training setup splits historic data into two sequential parts. The first part is used as input into TimesFM and the second part is used to train the MLP using the TimesFM forecast along with our GNN embeddings. A second set of TimesFM predictions using the full historic data is also generated for comparison. We evaluated performance using the unemployment rate across the US at the County level and Poverty Rate across the US at the postal code level. We present mean absolute percentage error (MAPE) in the table (lower is better performance). The results show that the PDFM embeddings are able to augment TimesFM estimates and exceed the supervised model performance for both variables.}
\label{table:forecasting}
\centering
\begin{tabular}{|c|c|c|}
\hline
\rowcolor[HTML]{BFCCE3} 
\multicolumn{3}{|c|}{\textbf{Mean Absolute Percentage Error [Relative Improvement]}} \\ \hline
\rowcolor[HTML]{BFCCE3} 
\textbf{Model} & \textbf{\begin{tabular}[c]{@{}c@{}}Unemployment Rate\\ (County)\end{tabular}} & \textbf{\begin{tabular}[c]{@{}c@{}}Poverty Rate\\ (Postal Code)\end{tabular}} \\ \hline
TimesFM (t) & 0.1186 & 0.3215 \\ \hline
ARIMA (t) & 0.1160 {\color[HTML]{3cb043}{[}-2\%{]}} & 0.2934 {\color[HTML]{3cb043}{[}-9\%{]}} \\ \hline
TimesFM (t-1) + PDFM Embedding & \textbf{0.1132} {\color[HTML]{3cb043}{[}-5\%{]}} & \textbf{0.2577} {\color[HTML]{3cb043}{[}-20\%{]}} \\ \hline
\end{tabular}
\end{table}

Our primary objective was to assess whether these embeddings could improve forecasts made for future time periods (six months for unemployment data and two years for poverty data). The results in Table \ref{table:forecasting} demonstrate a comparison between models that utilize PDFM embeddings and those that do not. The results in Table \ref{table:forecasting} show that for the mean absolute percentage error (MAPE) metric, the PDFM is able to improve on the TimesFM baseline performance and surpass the performance of ARIMA, suggesting that incorporating the PDFM can provide value without having to retrain existing forecasting models. We used a paired sample t-test to compare predictions from TimesFM + PDFM Embeddings and ARIMA, showing there to be a statistically significant difference (p-value < 0.025) when accounting for multiple comparison correction using \cite{dunn1961multiple} in both poverty and unemployment.

This improvement can be attributed to the ability of PDFM embeddings to capture valuable information about the relationships between different counties or postal codes. These embeddings encode not only the individual characteristics of each location but also the spatial dependencies and interactions that exist between them such as distances between regions or affinities from features such as aggregated search trends. By incorporating this additional layer of information, the forecasting models are able to make more informed predictions that account for the broader socio-economic context.

\begin{figure} 
    \centering
    \includegraphics[width=\textwidth]{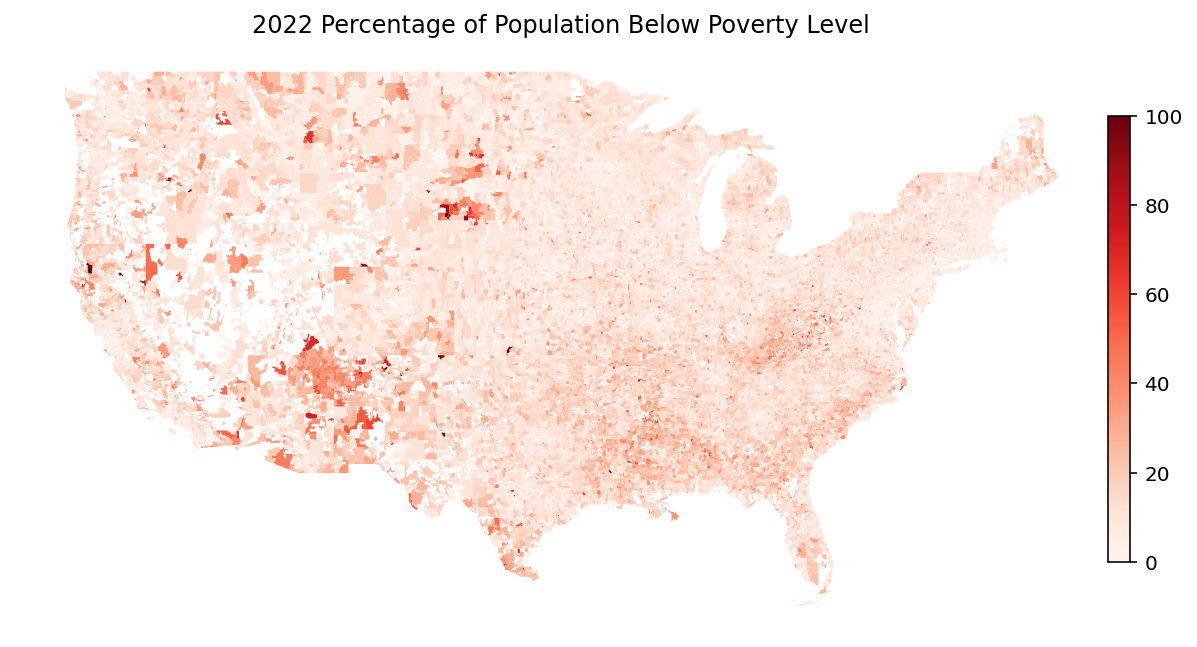}
    \caption{\footnotesize\textbf{Percentage of Population Below Poverty Line Used for Forecasting} Visualization of the poverty data from Data Commons for a single year at the postal code level in the United States used in the experiment to evaluate the benefit of embeddings extracted from the PDFM used as covariates for TimesFM. Demonstrating the high resolution of the data and how we are able to capitalize on that rich stratification by incorporating that knowledge through our PDFM embedding that is missing from univariate forecasting models.}
    \label{fig:2022_percentage_below_poverty_level}
\end{figure}

Figure \ref{fig:2022_percentage_below_poverty_level} is a visualization of the poverty data from Data Commons for a single year at the postal code level in the United States used in the experiment to evaluate the benefit of embeddings extracted from the PDFM used as covariates for TimesFM. Demonstrating the high resolution of the data and how we are able to capitalize on that rich stratification by incorporating that knowledge through our PDFM embedding that is missing from univariate forecasting models.

\section{Conclusion}

We have presented the PDFM, a flexible foundational model framework capable of addressing a wide range of geospatial challenges in the contiguous United States. By integrating diverse datasets, PDFM embeddings outperforms existing SoTA location-encoder approaches like SatCLIP and GeoCLIP across 27 health, socioeconomic, and environmental tasks, demonstrating superior capabilities in all 27 tasks for interpolation, and 25 tasks for extrapolation and super-resolution. Furthermore, we showcase how PDFM embeddings can enhance forecasting models like TimesFM, leading to improved predictions for crucial socioeconomic indicators like unemployment at the county level and poverty at the postal code level.  The adaptability of the PDFM to new tasks and its effectiveness in scenarios with limited data or varying output resolutions underscores its potential for widespread applications in research, social good initiatives, public and environmental health, and business sectors.

\section{Discussion}

Our results show that a foundation model with a rich and diverse set of datasets can broadly model a wide range of problems. Individually, signals like aggregated search trends perform best for predicting physical inactivity, while maps perform best at predicting night lights or population density, even surpassing remote sensing signals. However, combining multiple signals into one foundation model outperforms all individual signals. This unlocks the potential for our architecture to easily incorporate new signals and be applied to new domains and downstream tasks without having to hand craft features or design new model architectures.

We found that for interpolation and super-resolution, incorporating a signal such as SatCLIP has a clear benefit, but reduces performance for extrapolation. One aspect we intend exploring is how these signals can provide value as we create a global model. For example, because of comparable performance between weather and air quality with SatCLIP, we could rely on SatCLIP features in parts of the world where one or the other isn’t available spatially or temporally.

Our results show that for the elevation super-resolution task, IDW had the optimal performance compared to SatCLIP, GeoCLIP, or PDFM. Our explanation is that the change in elevation across the contiguous US is gradual and IDW is easily able to model this process, while other models may be confused by a combination of both built environment signals and features emphasizing human behaviors.

To apply the PDFM to different interpolation, extrapolation, and super-resolution problems we simply need some target label dataset (at varying resolution and availability) while the PDFM datasets, model architectures, and training setup can stay fixed. This could be built into a platform where researchers, organizations and business can easily adapt the framework to their needs without having to invest the people and traditional domain specific datasets and features necessary to model these downstream tasks accurately. This could enable scenarios where there is limited data from part of the country (for example, about disease prevalence) and PDFM is able to predict that data across the rest of the country or at a higher resolution.

We also showed how we are able to extract embeddings from PDFM and use them to complement existing forecasting models. In particular, we focused on the state-of-the-art forecasting foundation model, TimesFM, showing that the combination of TimesFM predictions and PDFM embeddings allowed us to exceed fully supervised forecasting performance. One advantage of using PDFM embeddings is that we are able to provide geospatial covariates that improve the forecast quality without having to fine tune either foundation model. This template can be used with other existing approaches where the embeddings from the PDFM can be treated as a fingerprint of a location to augment existing models that could include other forecasting approaches, existing geospatial classification and regression models, super-resolution methods, or Large Language Models (LLMs). 

Our approach has some limitations. First, the temporal granularity, particularly for graph inputs, is nonideal and not aligned between covariates. For example, aggregated search trends were obtained over July of 2022 while maps data were obtained from May of 2024. Our results are promising despite this temporal mismatch, though a better understanding of temporal alignment and its effects could boost performance even further. Similarly, the current PDFM approach produces static embeddings; adding a time dimension to the graph where message passing is done over time could be an interesting future direction when data are available. The selection of input data used in the PDFM is not exhaustive and each dataset provides unique value. Future research could explore other diverse datasets, including those derived from public social media, census data, location-specific webpages, sales statistics, traffic flows, and news articles.
Another limitation is the quality and source of our ground truth geospatial data. For example, the health target variables at the postal code and county level we used from CDC Places (\cite{cdcplaces2024}) were derived from a small area estimation model using existing census data. Ideally, there would be higher quality and higher resolution (both temporally and spatially) data to better evaluate the performance on these categories.
The PDFM incorporates both spatial proximity and feature similarity based graph edges. In future work we intend to explore more types of non-spatial edges. This would allow the nodes to learn through long distances that are not geographically connected but present similar patterns on some modalities. We also intend to explore new features that are available only to graph methods, such as edge only features. For example, going from Oakland to San Francisco involves a toll bridge only in one direction. Graphs can encode this asymmetric relationship that we have not fully explored with the labels we used. Further, datasets like aggregated mobility, capturing how people move across postal codes and counties, would bring interesting insights in epidemiological and health predictions via GNNs.

One obstacle to creating a foundation model for geographically structured data is the lacking availability of features and labels in different parts of the world, especially outside of the US and Europe. This opens up a novel research direction of operating in low-data regimes. Predictions made in such regimes could be accompanied with reliability estimates for those predictions based on the features and labels of neighboring nodes in the graph.

Our work focuses on aggregating our input signals at relatively large spatial scales (postal code and county), longer temporal periods (on the order of months), and choosing the most common aggregated search trends or maps categories without retaining the actual queries or categories in our features. We have shown this feature design is generally applicable but it also helps to preserve privacy. In future scenarios where the PDFM framework incorporates more sensitive signals mapping sparser populations, techniques such as differential privacy \cite{bavadekar2021google, kaplan2019differentially} will need to be applied to provide clear privacy guarantees.

\section{Methods}\label{methods}

\subsection{Datasets}

We collected and curated five general datasets to train the PDFM at the postal code level in the contiguous US (CONUS). All experiments and evaluations in this paper focus on the CONUS region.

\textbf{Aggregated Search Trends.} We computed the aggregate counts of the top 500 queries that were searched at least 20 times during July of 2022 in each postal code (\cite{sun2024community}). This resulted in over 1MM unique queries. We ranked the queries by their countrywide popularity, represented by the total number of postal codes that each query appeared in. We selected the top 1,000 queries from this list as a representation of aggregated search trends activity at the postal code level across the country. The lowest ranked or least common query appeared in over 700 postal codes. In each postal code we built our 1,000 dimensional feature vector with its respective frequencies for each query and scaled the vector to sum to 100. We designed these features to preserve privacy by ensuring we selected the most common queries. Each of the 1,000 queries occurred at least 14,000 times in the dataset and the query texts are ablated from subsequent analysis. The resulting dataset contained no personally identifiable information.

We did not have access to aggregated search trends in regions under 3 square-kilometers in area or with a population of less than 1,000 people, and we filtered out postal codes where there were no aggregated search trends for over 98\% of the queries. The resulting set of locations includes ~28,000 postal codes and covers over 95\% of the US population. County level data is computed as the unweighted mean of the values from the contained postal codes.

\textbf{Maps.} We selected the 1,192 most common point-of-interest categories from Google Maps during May of 2024 represented in at least 5\% of postal codes. Each category encompasses a broad range of point of interest locations - for example, the category “health care facility” includes children’s hospitals and university hospitals. We computed the total number of available facilities within the geospatial boundary to create a normalized 1,192 dimensional feature vector at the postal code and county levels.

\textbf{Busyness.} For each of the point of interest categories in our maps data, we computed a summary of the busyness of those categories by aggregating visits at all relevant locations over an entire month for each category. The resulting feature vector is normalized at the postal code and county level to account for population density. This feature is derived from data from the first half of 2024 and not temporally aligned with datasets from July of 2022.

\textbf{Weather \& Air Quality.} We collected weather and air quality readings and aggregated hourly data across July of 2022 summarized by mean, minimum, and maximum values. Finer grained daily readings were normalized across postal codes before being aggregated at the monthly level. The full list of values include mean sea level pressure, total cloud cover, 10 meter U wind, 10 meter V wind, temperature at a 2 meter height, dew point temperature at a 2 meter height, total sky direct short-wave (solar) radiation at surface, total precipitation rate, air quality index, carbon monoxide concentration, nitrogen dioxide concentration, ozone concentration, sulfur dioxide concentration, inhalable particulate matter (<10µm) concentration, and fine particulate matter (<2.5µm) concentration. 

\textbf{Remote Sensing.} We incorporated satellite imagery embeddings derived from the ViT16-L40 version of SatCLIP (\cite{klemmer2023satclip}) using the centroid of each postal code as the index to each embedding. The SatCLIP model aims to be a global, general-purpose geographic location encoder and summarizes 100,000 tiles of Sentinel-2 satellite imagery captured between January 1, 2021 and May 17, 2023.

\textbf{Temporal Alignment.} Only the aggregated search trends, weather, and air quality datasets are aligned temporally to July of 2022. We did experiment with using data sources aggregated web search trends from different months and found almost identical performance. Evaluating temporal consistency across modalities for the tasks presented here were out of scope for this work but could provide additional benefits when available.

\subsection{Model Architectures}

In this section, we describe the model architecture for our foundational model that generated embeddings to enable downstream tasks (phase 1), and the lightweight architectures of the models that took in the embeddings as inputs and were tuned to those downstream tasks (phase 2). 

\subsubsection{Graph Neural Network Foundation Model (Phase 1)}

\textbf{Graph Creation.} We construct a heterogeneous geospatial graph using county and postal codes as nodes, and edges via proximal relationships. The constructed geospatial graph has homogeneous node sets, representing both postal code and county nodes as the same type of node sets, and heterogeneous edge sets, with different types of edges connecting the nodes. The feature vectors from all the datasets are represented separately in the graph schema during the graph construction stage. The nodes in the graph are connected via two distinct kinds of edges, namely, proximity-based edges and relationship-based edges. This allows the postal codes and county nodes to gather additional information from postal codes or counties which are geographically close and similar in behavioral patterns (modeled using search similarity).

\textbf{Proximity- and Relationship- based Edges.} Counties and postal codes are connected to each other according to the geographical vicinity. Specifically, we create bidirectional edges between a postal code and another postal code or a county and another county if their centroids are within 100 miles distance of each other (see ZIP Code Distance Database). County and postal code nodes are connected to each other based on geographical relationship:  an edge exists if there is an overlapping geographical boundary. The relationship edges are based on the similarity between features such as aggregated search trends. 

\textbf{Subgraph Sampling.} We perform subgraph sampling to create subgraphs for training large-scale GNNs and adding stochasticity to the model as recommended by the GraphSAGE framework introduced in \cite{hamilton2017inductive}. For each geographical node (treated as the seed node), we sample both postal code and county nodes up to four hops away from the seed node. Specifically, we start from the seed node, traverse in a breadth-first fashion for each edge set, sample a fixed number of nodes in a weighted manner, and terminate when we’re four hops away from the seed node. This approach leads to a total number of subgraphs that is equal to the total number of postal code and county nodes. Each of these subgraphs represents the local neighborhood associated with its corresponding seed node. For modeling, each subgraph and the corresponding label serve as a single data point for training and validation purposes. 

\textbf{Pre-Processing.} Column-wise standardization is applied to all features, and the extreme ends of the feature value ranges are compressed via clipping.  For every node in the input subgraphs, the feature vectors from the source datasets (e.g., search trends, maps, etc.) are concatenated to form a single feature vector. We add a single hidden layer on top of it to create a feature vector of fixed length, with a Gaussian Error linear unit (GeLU) for non-linearity. This defines the initial hidden state of the GNN before message passing occurs. 

\textbf{Modeling and Training Details.} We employ GraphSAGE, an inductive approach, to learn the node embeddings by leveraging the node feature information. The GraphSAGE implementation learns a function that generates embeddings by aggregating from the local neighborhood (already sampled in the subgraph creation). For the aggregation architecture, we use the pooling architecture presented in GraphSAGE, where node states from the neighborhood nodes are fed through a fully-connected layer with ReLU transformation. The transformed old states and neighborhood node states are further aggregated with an element-wise sum operation. We use a single round of message passing facilitated by the GraphSAGE architecture. After the GNN layer, we add a single linear layer of dimension 330 to encode the node-level representations into compressed embeddings.

The model is trained via self-supervised learning, where input features are also used as labels. The embedding layer is partitioned into three separate parts, each dedicated to encoding and reconstructing a subset of input features based on the source type. Dimensions 0-127 are used to reconstruct the Aggregated Search Trends, dimensions 128-255 are used to reconstruct the Maps and Busyness features, and dimensions 256-329 are used for Weather \& Air Quality. The gradients are propagated into the full network via Huber loss using the Adam optimizer with a cosine-decayed learning rate. 

\begin{figure} 
    \centering
    \includegraphics[width=\textwidth]{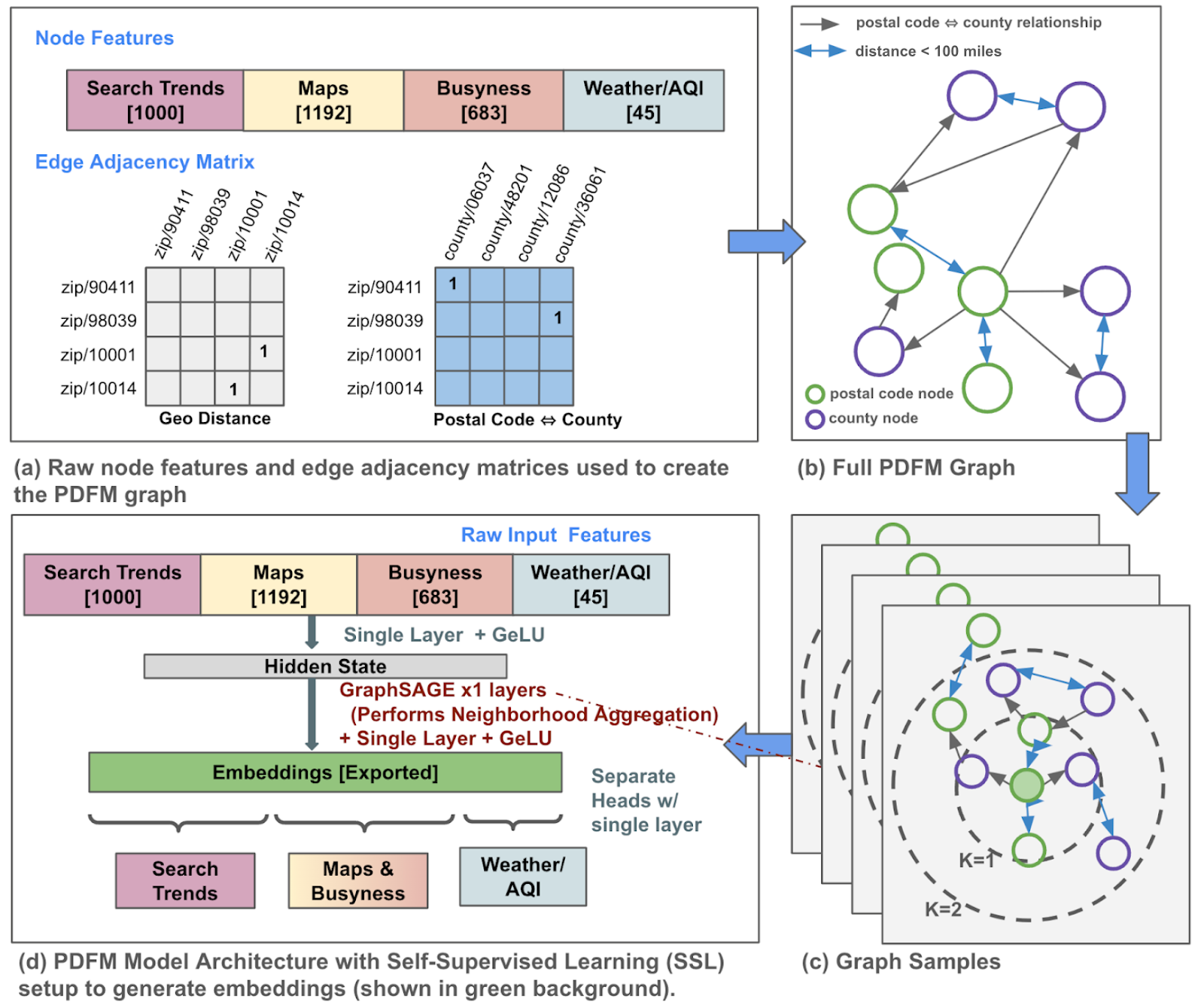}
    \caption{\footnotesize\textbf{Graph Neural Network Architecture Diagram} (a) Presents the raw node features and edge adjacency matrix which is used to generate the (b) full PDFM graph. (c) Subgraph sampling was performed for each postal code node as the seed to generate multiple subgraphs. (d) Overview of the GNN architecture and layers including GraphSAGE showing neighborhood aggregation in the sampled subgraphs, with self-supervised setup to reconstruct input features for generating embeddings. }
    \label{fig:graph_architecture}
\end{figure}

\textbf{Hyperparameter Tuning.} We uniformly sampled 20\% of the seed nodes (including both counties and zip codes) into a validation set for tuning. Since we trained in a self-supervised manner, only the seed node features (acting as labels) were affected by this split, while edges and other associated node data were kept for the full graph. The hyperparameters for the PDFM were tuned on this validation split of the data, then used to train a final model over all nodes. The hyperparameters that were tuned included dropout rate, the sizes of node embeddings, numbers of GraphSAGE hidden units and layers, embedding sizes, regularization, and learning rate.

\subsubsection{Downstream Models (Phase 2)}

To assess the predictive power and ease of use of our embeddings, we compared the performance of three popular types of models over three geospatial tasks and 27 labels. The results are presented in Figure \ref{fig:phase2_model_comparison}.  The models included in this comparison are linear regression, gradient boosted decision trees (GBDT), and multilayer perceptrons (MLP).  The results show that using a simple linear regression model with PDFM embeddings works fairly well, and more complex models improve performance except for in the case of super-resolution. For super-resolution, there are 10x less data points for training than for the other tasks, which makes it more difficult to avoid overfitting with more complex models. 

\textbf{Linear Regression.} For each task and label combination, we trained a Ridge regression model with default configurations from the popular Scikit-learn library (\cite{fabian2011scikit}).

\textbf{Multilayer Perceptron (MLP).}  MLP is a type of feedforward neural network suitable for modeling nonlinear relationships between variables. For each task and label combination, we used the TensorFlow (\cite{abadi2016tensorflow}) library to train a three layer MLP with rectified linear (ReLU) activation and a 20\% dropout rate, using the Adam optimizer with an initial learning rate of 0.005 and cosine decay over 40 epochs. The layer dimensions are 512, 256, and 128.

\textbf{Gradient Boosted Decision Trees (GBDT).} GBDT is a common state of the art method for modeling tabular data (\cite{shwartz2022tabular, grinsztajn2022why}). We used the GBDT implementation from the popular LightGBM (Ke, 2017) library with lightly tuned hyperparameters: up to 3000 trees with a maximum of 31 leaves per tree and a minimum of 40 samples per leaf, trained with a learning rate of 0.02. Each model trained in a few minutes on a single CPU and typically performed better than linear regression.

\textbf{Single Modality Evaluation.} In addition to training the above models using the full PDFM embedding vectors, we also repeated all experiments using the subsets of the embedding dimensions dedicated to encoding each category of the source data, namely Aggregated Search Trends (dimensions 0-127),  Maps and Busyness (dimensions 128-255), and Weather \& Air Quality (dimensions 256-329). The per-modality metrics are included in the results section.

\subsection{Benchmarks}

\subsubsection{Problem Definition}

We evaluated our PDFM on three geospatial problems: interpolation, extrapolation, and super-resolution.

\textbf{Interpolation}

Missing data is a common issue with surveys such as the census that might miss certain demographics, locations, or answers (\cite{brick1996handling}). The goal of imputation via spatial interpolation is to generate the most likely data to substitute those missing values. 

\begin{figure} 
    \centering
    \includegraphics[width=\textwidth]{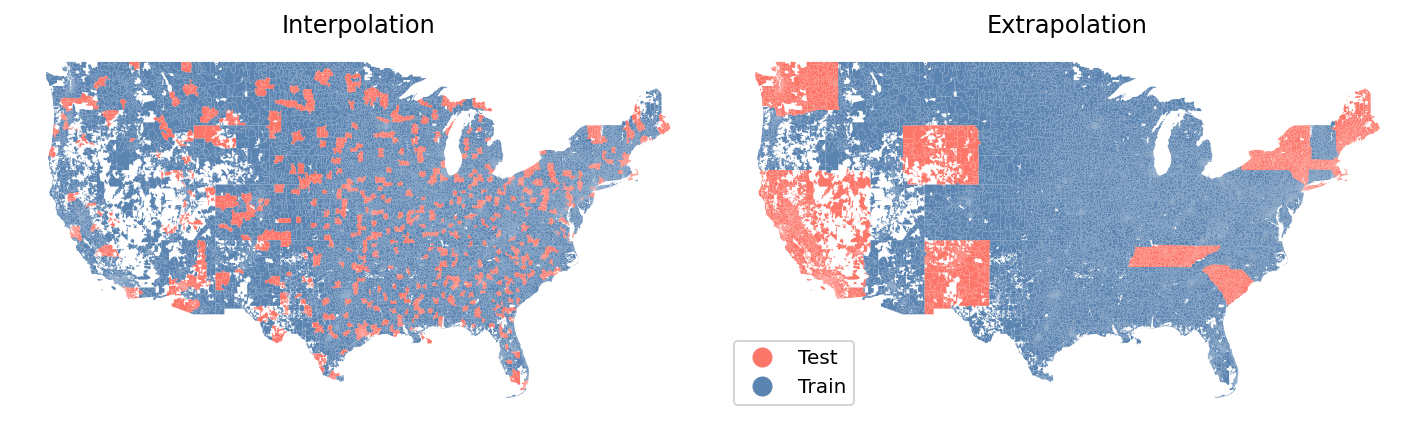}
    \caption{\footnotesize\textbf{Geographic Training Data Splits.} Map showing the dataset splits, left for interpolation and right for extrapolation. Red areas are counties where all contained postal codes are in the holdout set, blue areas are used for training and validation.}
    \label{fig:dataset_splits}
\end{figure}

First, we created a one-to-one mapping from each postal code to a single county based on the largest land area overlap. Then we randomly selected 20\% of the counties and all the postal codes contained within them as the holdout test set.  This approach creates more spatial separation between the training and test data than uniformly selecting postal codes for the test set, and mimics the more difficult types of real world missing-data situations more closely. Figure \ref{fig:dataset_splits} shows the different areas that are used for training and testing. 

We used a 20\% subset of the training set for validation during model development, and the holdout set is only used for final testing and visualization.

\textbf{Extrapolation}

For the extrapolation task we randomly selected 20\% of the states and all the postal codes contained within them (based on the postal code to county mapping described earlier) as the holdout test set, and used the rest of the states for training and validation. 

\textbf{Super-resolution}

For the super-resolution task, we used only county-level labels for training, 20\% of the interpolation training set postal codes for validation, and the interpolation test set postal codes as the holdout test set.

\subsubsection{Benchmark Datasets}\label{sec:benchmark_datasets}

We used an existing general geospatial benchmark by (\cite{sun2024community}),  retrieved from \cite{datacommons2024} and the Earth Engine catalog (\cite{gorelick2017google}). Our health tasks are all of the 2022 health related metrics from CDC Places that were available in Data Commons at the postal code resolution. We selected all socioeconomic and environmental tasks that were used by \cite{rolf2021generalizable} and available in Data Commons or the Earth Engine Catalog. These included income, home value, night lights, tree cover, and elevation. \cite{klemmer2023satclip} did a comparison of median income, home value, elevation, and population across a range of geolocation encoders. They found SatCLIP and GeoCLIP to be the optimal encodings. We evaluated both SatCLIP and GeoCLIP across all of our variables in our experiments.

\cite{tkachenko2017google} studied diabetes by correlating specific queries with candidate risk variables. \cite{cesare2019use} used search trends related to physical activity and diet along with social media, demographics, and built environment variables to predict 2013 county level obesity prevalence across the US using a linear model with $R^2$ of 0.624 and 0.55 for females and males respectively. Smoking, binge drinking, stroke, coronary heart disease, obesity, diabetes, and poor physical health in 208 Designated Market Areas (DMAs) across the US by selecting categories of search trends related to these tasks (\cite{jaidka2021information}). We build on these examples showing how we are able to improve performance by incorporating complementary datasets for these types of health tasks.

In addition to the 27 spatial tasks, we selected both unemployment available at the county level and poverty available at the postal code level because of their high temporal resolution. Both were retrieved from Data Commons. See Section \ref{sec:applications} for additional statistics about these two tasks in the context of forecasting.

The full list of 27 spatial and 2 forecasting tasks includes the following.
\begin{itemize}
\item \textbf{21 Health:} High Cholesterol, Physical Health Not Good, Stroke, Binge Drinking, Physical Inactivity, Received Annual Checkup, Cancer (Excl. Skin Cancer), Diabetes, Mental Health Not Good, Coronary Heart Disease, High Blood Pressure, Received Cholesterol Screening, Received Dental Visit, Asthma, Chronic Kidney Disease, Arthritis, Chronic Obstructive Pulmonary Disease, High Blood Pressure (Medicated), Obesity, Sleep Less Than 7 Hours, Smoking
\item \textbf{6 Socioeconomic}
    \begin{itemize}
    \item \textbf{4 Spatial}: Median Income Household, Median Home Value, Night Lights, Population Density
    \item \textbf{2 Forecasting}: Unemployment, Poverty
    \end{itemize}
\item \textbf{2 Environmental:} Tree Cover, Elevation
\end{itemize} 

\begin{figure} 
    \centering
    \includegraphics[width=\textwidth]{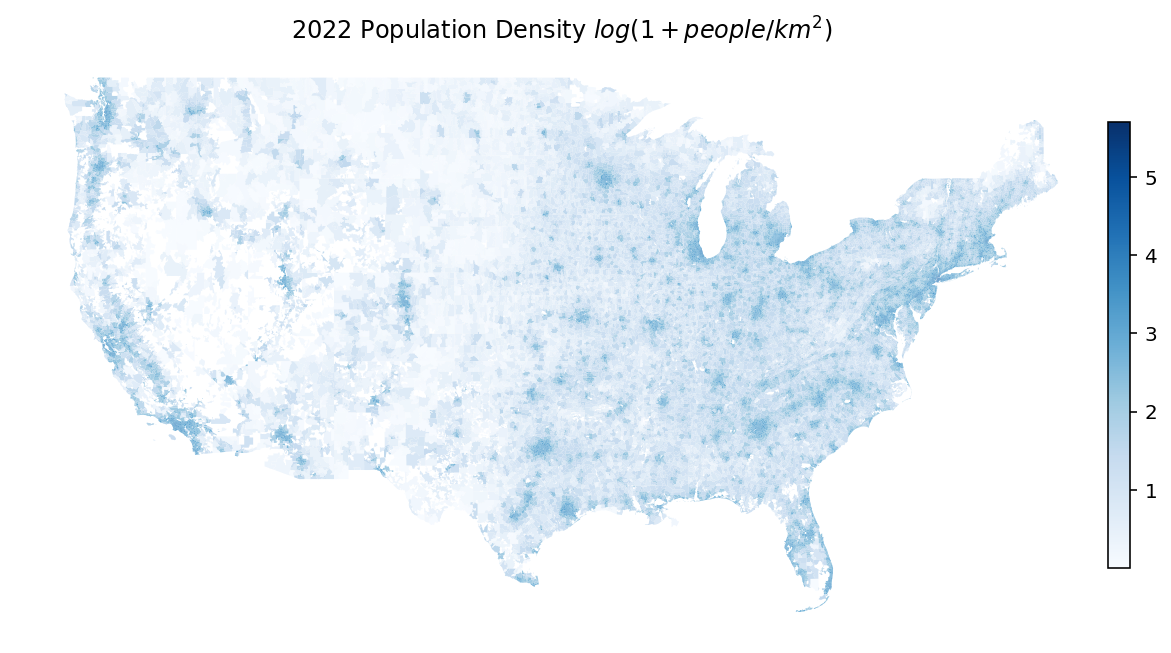}
    \caption{\footnotesize\textbf{Population Density Visualization in CONUS} We show a visualization of relatively high resolution population density from 2022 used in our experiments for model evaluation.}
    \label{fig:2022_population_density}
\end{figure}

See Figure \ref{fig:2022_population_density} for a visualization of the population density data used in our benchmark analysis. The data for each downstream task was available at both the postal code and county level for the contiguous US.

\subsubsection{Comparison Approach}\label{comparison_approach}

\subsubsubsection{Geographic Location Encoders}

We selected two geographic encoders that showed strong performance on income, housing prices, elevation, and population density in previous experiments (\cite{klemmer2023satclip}). We evaluate the same four tasks in our experiments in addition to 23 other tasks as part of our benchmark.

\textit{Satellite Contrastive Location-Image Pretraining (SatCLIP)} by \cite{klemmer2023satclip} is a geographic encoder based on satellite imagery. SatCLIP outperforms a previous version called MOSAIKS (\cite{rolf2021generalizable}) along with a technique using contrastive self-supervision (CSP) (\cite{mai2023csp}), that was trained on species and satellite imagery.

\textit{GeoCLIP} (\cite{cepeda2024geoclip}) is a geographic encoder trained with 4.72 million Flickr images. GeoCLIP often outperforms SatCLIP on income and housing benchmarks.

\subsubsubsection{Interpolation Methods}

We evaluated three state-of-the-art interpolation approaches as a baseline for the 27 downstream tasks of our foundation model. These techniques were implemented by learning a model on the training split of the downstream tasks or ground truth without incorporating any outside or additional information.

\textit{Inverse Distance Weighting (IDW)} (\cite{shepard1968two}) is a technique that is able to interpolate by learning a weighted average of known points. In our case IDW operates by learning only on the latitude and longitude coordinates along with the downstream task labels within our training and validation splits without any additional data. These results are presented alongside our foundation model results.

\textit{Radial Basis Function (RBF) Interpolation} (\cite{fasshauer2007meshfree}) uses an RBF centered at a target location in combination with a polynomial function to fill in missing values. These results are not presented in this paper because of significantly lower performance compared with IDW.

\textit{Gaussian Process Regression} (\cite{chung2019supplement}) incorporates the spatial correlation of sampled points when predicting missing values. These results are not presented in this paper because of significantly lower performance compared with IDW.

The results from our experiments showed that IDW consistently performed the best across all the tasks and we use that as our state-of-the-art interpolation method when presenting results. \textit{We do not present results using RBF or Gaussian Process Regression because they were both significantly lower performing methods.}

\subsection{Application of PDFM Embeddings}
\label{sec:applications}

The PDFM architecture allows us to easily extract embeddings from the network that can be incorporated into downstream models. The embeddings are meant to be a compact representation of the most salient features across all of the input datasets and tuned in a way to be generally useful for downstream tasks. Embeddings extracted after the message passing layer enable the model to capture any n-hop neighborhood information within the embedding itself. In our experiments, we extracted embeddings from the final hidden layer ( after a round of message passing through GraphSAGE) to ensure the embeddings contained the full context of the local neighborhood. 

\textbf{Forecasting.} We evaluated the utility of the PDFM embeddings by augmenting an existing forecasting model. We selected TimesFM (Time Series Foundation Model) (\cite{das2023decoder}). TimesFM is a state-of-the-art pre-trained decoder attention model utilizing patched inputs from a large real-world and synthetic time-series corpus. It required no training when applied to a new forecasting problem (although incorporating our PDFM embeddings does require training a simple adapter model, such as the multilayer perceptron (MLP), described below in our experimental setup). As a benchmark and way to calibrate our results, we also incorporated a traditional supervised statistical forecasting approach, ARIMA (AutoRegressive Integrated Moving Average) (\cite{box1976time}). ARIMA leverages past values of the input data and past forecast errors to capture trends and account for noise within the data.

\textbf{Experimental Setup.} To evaluate whether augmenting TimesFM with our embeddings provides a benefit in forecasting performance, we implemented a two-stage process to incorporate embeddings extracted from PDFM as covariates to adjust for errors in the TimesFM predictions. The first stage involves utilizing the base model to get the baseline prediction, meaning we take historic data and split it into three sequential parts. The first part is used to generate a forecast from TimesFM for the second part. The second stage introduces additional information through a Multi-Layer Perceptron (MLP) with two hidden layers. The input consists of the predictions generated by the TimesFM predicting the second part of the historic data along with PDFM embeddings. The ground truth values over the second part of historic data serve as the target variable for this MLP training stage. Finally we evaluate the performance of combining TimesFM and PDFM embeddings using the trained MLP on the third part of historical data, and compare it to a new set of TimesFM predictions based on both part one and part two of the data.

\textbf{Datasets.} We utilized two datasets at different spatial (postal code and county level) resolutions for our experiments. We excluded counties or postal codes with missing data on a per task basis. This ensures the models are trained and evaluated on data with similar sampling to better evaluate performance.
The unemployment dataset encompasses monthly unemployment figures for each county, spanning January of 1990 to March of 2024. Because most of the datasets in the PDFM embeddings are from July of 2022, we leverage data from January of 1990 to June of 2022 as part one and July of 2022 as part two of the historical context for our forecasting models. This context window is then utilized to generate predictions for the subsequent six months, from August of 2022 to January of 2023. 
The poverty dataset (see Figure \ref{fig:2022_percentage_below_poverty_level} for example visualization of the data) includes yearly poverty counts and population figures for each postal code, covering the period from 2011 to 2022. We calculate the poverty rate for each postal code by dividing the poverty count by the corresponding population count. Additionally, to ensure data quality, we filter the dataset to retain only postal codes with poverty rates over 0.05. Data from 2011 to 2019 serves as part one and 2020 as part two of the historical context for predicting poverty rates in 2021 and 2022. 

\textbf{Training Splits.} For unemployment, we utilized the predictions and ground truth for July of 2022 in all counties for training, and the predictions for January of 2023 for testing. Similarly, for the poverty data, predictions for 2021 in all postal codes are employed for training the model, while predictions for 2022 are used for testing.
\clearpage

\section{Data and Code Availability}

We offer a program for academic and non-commercial researchers to access PDFM embeddings. To learn more, please email \href{mailto:pdfm-embeddings@google.com}{pdfm-embeddings@google.com}. Once access is granted, public code samples for using the PDFM embeddings and creating the benchmarks can be found at \url{https://github.com/google-research/population-dynamics}.

\section{Competing Interests}
This study was funded by Alphabet Inc and/or a subsidiary thereof (`Alphabet'). All authors are (or were) employees of Alphabet and may own stock as part of the standard compensation package.

\clearpage
\bibliography{main}

\clearpage
\appendix
\section*{Appendix}
\setcounter{table}{0}
\setcounter{figure}{0}
\renewcommand{\thetable}{A\arabic{table}}
\renewcommand{\thefigure}{A\arabic{figure}}
We provide additional details in Table \ref{table:interpolation_full} for the interpolation task, adding the performance metrics for PDFM with SatCLIP.  In 20 out of 27 tasks, concatenating PDFM and SatCLIP embeddings achieved the best performance while in the other 7 tasks it was on par with PDFM embeddings alone. Averaged over all 27 variables, PDF without SatCLIP achieves an R2 of 0.75, and adding SatCLIP improves it to 0.77. The integration of SatCLIP seems to enhance the models' ability to capture spatial patterns and relationships relevant to a wide array of health, socioeconomic, and environmental factors. While the performance gains vary across specific tasks, the consistent trend underscores the value of incorporating multimodal data, especially satellite imagery, in modeling complex real-world phenomena. 

\begin{table}[!hbt]
\fontsize{10pt}{10pt}\selectfont
\caption{\textbf{Interpolation $R^2$ with optimal bolded (Higher $R^2$ is Better).} We present our full set of experiments for interpolation across all 27 downstream tasks including health, socioeconomic, and environment categories using the $R^2$ metric (a higher value shows the respective model better accounts for the variance in the groundtruth target variable labels). These experiments compare inverse distance weighted interpolation (IDW), SatCLIP embeddings, GeoCLIP embeddings, our PDFM embeddings and its subcomponents (Weather \& Air Quality, Aggregated Search Trends, Maps, and Busyness), as well as a concatenation of PDFM and SatCLIP embeddings. This table indicates the model column with the optimal $R^2$ performance in bold and higher performance with darker green.}
\label{table:interpolation_full}
\begin{tabular}{lrrrrrrrr}
 & IDW & SatCLIP & GeoCLIP & \makecell{Weather \\ \& AQ} & Trends & Maps & PDFM & \makecell{PDFM w/ \\ SatCLIP} \\
HighCholesterol & {\cellcolor[HTML]{F7FCF5}} \color[HTML]{000000} 0.24 & {\cellcolor[HTML]{F6FCF4}} \color[HTML]{000000} 0.25 & {\cellcolor[HTML]{EAF7E6}} \color[HTML]{000000} 0.36 & {\cellcolor[HTML]{ECF8E8}} \color[HTML]{000000} 0.34 & {\cellcolor[HTML]{D1EDCB}} \color[HTML]{000000} 0.51 & {\cellcolor[HTML]{C1E6BA}} \color[HTML]{000000} 0.59 & {\cellcolor[HTML]{B2E0AC}} \color[HTML]{000000} 0.65 & \bfseries {\cellcolor[HTML]{AEDEA7}} \color[HTML]{000000} 0.67 \\
PhysicalHealthNotGood & {\cellcolor[HTML]{F7FCF5}} \color[HTML]{000000} 0.41 & {\cellcolor[HTML]{F6FCF4}} \color[HTML]{000000} 0.42 & {\cellcolor[HTML]{EBF7E7}} \color[HTML]{000000} 0.51 & {\cellcolor[HTML]{EBF7E7}} \color[HTML]{000000} 0.51 & {\cellcolor[HTML]{C1E6BA}} \color[HTML]{000000} 0.72 & {\cellcolor[HTML]{C9EAC2}} \color[HTML]{000000} 0.69 & {\cellcolor[HTML]{B4E1AD}} \color[HTML]{000000} 0.77 & \bfseries {\cellcolor[HTML]{AEDEA7}} \color[HTML]{000000} 0.79 \\
Stroke & {\cellcolor[HTML]{F7FCF5}} \color[HTML]{000000} 0.27 & {\cellcolor[HTML]{F4FBF1}} \color[HTML]{000000} 0.30 & {\cellcolor[HTML]{EDF8E9}} \color[HTML]{000000} 0.36 & {\cellcolor[HTML]{EBF7E7}} \color[HTML]{000000} 0.38 & {\cellcolor[HTML]{C4E8BD}} \color[HTML]{000000} 0.59 & {\cellcolor[HTML]{C2E7BB}} \color[HTML]{000000} 0.60 & {\cellcolor[HTML]{B0DFAA}} \color[HTML]{000000} 0.67 & \bfseries {\cellcolor[HTML]{AEDEA7}} \color[HTML]{000000} 0.68 \\
BingeDrinking & {\cellcolor[HTML]{F7FCF5}} \color[HTML]{000000} 0.39 & {\cellcolor[HTML]{EFF9EC}} \color[HTML]{000000} 0.43 & {\cellcolor[HTML]{F3FAF0}} \color[HTML]{000000} 0.41 & {\cellcolor[HTML]{F3FAF0}} \color[HTML]{000000} 0.41 & {\cellcolor[HTML]{D1EDCB}} \color[HTML]{000000} 0.54 & {\cellcolor[HTML]{E7F6E3}} \color[HTML]{000000} 0.47 & {\cellcolor[HTML]{B6E2AF}} \color[HTML]{000000} 0.61 & \bfseries {\cellcolor[HTML]{AEDEA7}} \color[HTML]{000000} 0.63 \\
PhysicalInactivity & {\cellcolor[HTML]{F7FCF5}} \color[HTML]{000000} 0.42 & {\cellcolor[HTML]{F2FAEF}} \color[HTML]{000000} 0.46 & {\cellcolor[HTML]{ECF8E8}} \color[HTML]{000000} 0.51 & {\cellcolor[HTML]{EAF7E6}} \color[HTML]{000000} 0.52 & {\cellcolor[HTML]{BCE4B5}} \color[HTML]{000000} 0.74 & {\cellcolor[HTML]{C1E6BA}} \color[HTML]{000000} 0.72 & \bfseries {\cellcolor[HTML]{AEDEA7}} \color[HTML]{000000} 0.79 & \bfseries {\cellcolor[HTML]{AEDEA7}} \color[HTML]{000000} 0.79 \\
ReceivedAnnualCheckup & {\cellcolor[HTML]{F7FCF5}} \color[HTML]{000000} 0.67 & {\cellcolor[HTML]{EDF8E9}} \color[HTML]{000000} 0.70 & {\cellcolor[HTML]{F0F9ED}} \color[HTML]{000000} 0.69 & {\cellcolor[HTML]{F0F9ED}} \color[HTML]{000000} 0.69 & {\cellcolor[HTML]{D6EFD0}} \color[HTML]{000000} 0.75 & {\cellcolor[HTML]{F4FBF1}} \color[HTML]{000000} 0.68 & {\cellcolor[HTML]{BCE4B5}} \color[HTML]{000000} 0.79 & \bfseries {\cellcolor[HTML]{AEDEA7}} \color[HTML]{000000} 0.81 \\
Cancer & {\cellcolor[HTML]{F7FCF5}} \color[HTML]{000000} 0.25 & {\cellcolor[HTML]{F5FBF2}} \color[HTML]{000000} 0.27 & {\cellcolor[HTML]{EAF7E6}} \color[HTML]{000000} 0.36 & {\cellcolor[HTML]{ECF8E8}} \color[HTML]{000000} 0.34 & {\cellcolor[HTML]{C2E7BB}} \color[HTML]{000000} 0.57 & {\cellcolor[HTML]{C0E6B9}} \color[HTML]{000000} 0.58 & \bfseries {\cellcolor[HTML]{AEDEA7}} \color[HTML]{000000} 0.65 & \bfseries {\cellcolor[HTML]{AEDEA7}} \color[HTML]{000000} 0.65 \\
Diabetes & {\cellcolor[HTML]{F7FCF5}} \color[HTML]{000000} 0.31 & {\cellcolor[HTML]{F2FAF0}} \color[HTML]{000000} 0.35 & {\cellcolor[HTML]{EDF8E9}} \color[HTML]{000000} 0.40 & {\cellcolor[HTML]{EBF7E7}} \color[HTML]{000000} 0.42 & {\cellcolor[HTML]{C3E7BC}} \color[HTML]{000000} 0.64 & {\cellcolor[HTML]{C1E6BA}} \color[HTML]{000000} 0.65 & {\cellcolor[HTML]{B0DFAA}} \color[HTML]{000000} 0.72 & \bfseries {\cellcolor[HTML]{AEDEA7}} \color[HTML]{000000} 0.73 \\
MentalHealthNotGood & {\cellcolor[HTML]{F7FCF5}} \color[HTML]{000000} 0.29 & {\cellcolor[HTML]{F0F9ED}} \color[HTML]{000000} 0.36 & {\cellcolor[HTML]{E7F6E2}} \color[HTML]{000000} 0.45 & {\cellcolor[HTML]{E9F7E5}} \color[HTML]{000000} 0.42 & {\cellcolor[HTML]{C2E7BB}} \color[HTML]{000000} 0.66 & {\cellcolor[HTML]{C4E8BD}} \color[HTML]{000000} 0.65 & {\cellcolor[HTML]{B0DFAA}} \color[HTML]{000000} 0.74 & \bfseries {\cellcolor[HTML]{AEDEA7}} \color[HTML]{000000} 0.75 \\
CoronaryHeartDisease & {\cellcolor[HTML]{F7FCF5}} \color[HTML]{000000} 0.40 & {\cellcolor[HTML]{F7FCF5}} \color[HTML]{000000} 0.40 & {\cellcolor[HTML]{F0F9ED}} \color[HTML]{000000} 0.44 & {\cellcolor[HTML]{F2FAEF}} \color[HTML]{000000} 0.43 & {\cellcolor[HTML]{CBEAC4}} \color[HTML]{000000} 0.59 & {\cellcolor[HTML]{C8E9C1}} \color[HTML]{000000} 0.60 & {\cellcolor[HTML]{B1E0AB}} \color[HTML]{000000} 0.66 & \bfseries {\cellcolor[HTML]{AEDEA7}} \color[HTML]{000000} 0.67 \\
HighBloodPressure & {\cellcolor[HTML]{F7FCF5}} \color[HTML]{000000} 0.44 & {\cellcolor[HTML]{F6FCF4}} \color[HTML]{000000} 0.45 & {\cellcolor[HTML]{EEF8EA}} \color[HTML]{000000} 0.50 & {\cellcolor[HTML]{EEF8EA}} \color[HTML]{000000} 0.50 & {\cellcolor[HTML]{C8E9C1}} \color[HTML]{000000} 0.67 & {\cellcolor[HTML]{C2E7BB}} \color[HTML]{000000} 0.69 & \bfseries {\cellcolor[HTML]{AEDEA7}} \color[HTML]{000000} 0.75 & \bfseries {\cellcolor[HTML]{AEDEA7}} \color[HTML]{000000} 0.75 \\
CholesterolScreening & {\cellcolor[HTML]{F5FBF3}} \color[HTML]{000000} 0.36 & {\cellcolor[HTML]{EDF8EA}} \color[HTML]{000000} 0.43 & {\cellcolor[HTML]{EAF7E6}} \color[HTML]{000000} 0.46 & {\cellcolor[HTML]{F7FCF5}} \color[HTML]{000000} 0.34 & {\cellcolor[HTML]{D1EDCB}} \color[HTML]{000000} 0.61 & {\cellcolor[HTML]{DEF2D9}} \color[HTML]{000000} 0.54 & {\cellcolor[HTML]{BAE3B3}} \color[HTML]{000000} 0.72 & \bfseries {\cellcolor[HTML]{AEDEA7}} \color[HTML]{000000} 0.77 \\
DentalVisit & {\cellcolor[HTML]{F7FCF5}} \color[HTML]{000000} 0.38 & {\cellcolor[HTML]{EFF9EB}} \color[HTML]{000000} 0.46 & {\cellcolor[HTML]{E8F6E3}} \color[HTML]{000000} 0.52 & {\cellcolor[HTML]{E5F5E1}} \color[HTML]{000000} 0.54 & {\cellcolor[HTML]{B7E2B1}} \color[HTML]{000000} 0.77 & {\cellcolor[HTML]{C1E6BA}} \color[HTML]{000000} 0.73 & {\cellcolor[HTML]{B0DFAA}} \color[HTML]{000000} 0.80 & \bfseries {\cellcolor[HTML]{AEDEA7}} \color[HTML]{000000} 0.81 \\
Asthma & {\cellcolor[HTML]{F7FCF5}} \color[HTML]{000000} 0.38 & {\cellcolor[HTML]{F1FAEE}} \color[HTML]{000000} 0.43 & {\cellcolor[HTML]{EBF7E7}} \color[HTML]{000000} 0.48 & {\cellcolor[HTML]{E9F7E5}} \color[HTML]{000000} 0.49 & {\cellcolor[HTML]{C3E7BC}} \color[HTML]{000000} 0.69 & {\cellcolor[HTML]{D2EDCC}} \color[HTML]{000000} 0.62 & {\cellcolor[HTML]{B4E1AD}} \color[HTML]{000000} 0.75 & \bfseries {\cellcolor[HTML]{AEDEA7}} \color[HTML]{000000} 0.77 \\
ChronicKidneyDisease & {\cellcolor[HTML]{F7FCF5}} \color[HTML]{000000} 0.23 & {\cellcolor[HTML]{F5FBF2}} \color[HTML]{000000} 0.25 & {\cellcolor[HTML]{ECF8E8}} \color[HTML]{000000} 0.32 & {\cellcolor[HTML]{EDF8EA}} \color[HTML]{000000} 0.31 & {\cellcolor[HTML]{CAEAC3}} \color[HTML]{000000} 0.51 & {\cellcolor[HTML]{C3E7BC}} \color[HTML]{000000} 0.54 & {\cellcolor[HTML]{B0DFAA}} \color[HTML]{000000} 0.61 & \bfseries {\cellcolor[HTML]{AEDEA7}} \color[HTML]{000000} 0.62 \\
Arthritis & {\cellcolor[HTML]{F7FCF5}} \color[HTML]{000000} 0.45 & {\cellcolor[HTML]{F6FCF4}} \color[HTML]{000000} 0.46 & {\cellcolor[HTML]{EDF8EA}} \color[HTML]{000000} 0.51 & {\cellcolor[HTML]{EFF9EC}} \color[HTML]{000000} 0.50 & {\cellcolor[HTML]{CAEAC3}} \color[HTML]{000000} 0.66 & {\cellcolor[HTML]{C7E9C0}} \color[HTML]{000000} 0.67 & {\cellcolor[HTML]{B1E0AB}} \color[HTML]{000000} 0.73 & \bfseries {\cellcolor[HTML]{AEDEA7}} \color[HTML]{000000} 0.74 \\
COPD & {\cellcolor[HTML]{F7FCF5}} \color[HTML]{000000} 0.50 & {\cellcolor[HTML]{F5FBF3}} \color[HTML]{000000} 0.51 & {\cellcolor[HTML]{EFF9EB}} \color[HTML]{000000} 0.55 & {\cellcolor[HTML]{EFF9EB}} \color[HTML]{000000} 0.55 & {\cellcolor[HTML]{C8E9C1}} \color[HTML]{000000} 0.70 & {\cellcolor[HTML]{CEECC8}} \color[HTML]{000000} 0.68 & {\cellcolor[HTML]{B1E0AB}} \color[HTML]{000000} 0.76 & \bfseries {\cellcolor[HTML]{AEDEA7}} \color[HTML]{000000} 0.77 \\
BPMed & {\cellcolor[HTML]{F7FCF5}} \color[HTML]{000000} 0.16 & {\cellcolor[HTML]{F4FBF2}} \color[HTML]{000000} 0.20 & {\cellcolor[HTML]{EDF8EA}} \color[HTML]{000000} 0.28 & {\cellcolor[HTML]{EEF8EA}} \color[HTML]{000000} 0.27 & {\cellcolor[HTML]{DCF2D7}} \color[HTML]{000000} 0.44 & {\cellcolor[HTML]{B4E1AD}} \color[HTML]{000000} 0.70 & \bfseries {\cellcolor[HTML]{AEDEA7}} \color[HTML]{000000} 0.73 & \bfseries {\cellcolor[HTML]{AEDEA7}} \color[HTML]{000000} 0.73 \\
Obesity & {\cellcolor[HTML]{F7FCF5}} \color[HTML]{000000} 0.48 & {\cellcolor[HTML]{ECF8E8}} \color[HTML]{000000} 0.55 & {\cellcolor[HTML]{E4F5DF}} \color[HTML]{000000} 0.60 & {\cellcolor[HTML]{E9F7E5}} \color[HTML]{000000} 0.57 & {\cellcolor[HTML]{BBE4B4}} \color[HTML]{000000} 0.75 & {\cellcolor[HTML]{C2E7BB}} \color[HTML]{000000} 0.73 & \bfseries {\cellcolor[HTML]{AEDEA7}} \color[HTML]{000000} 0.79 & \bfseries {\cellcolor[HTML]{AEDEA7}} \color[HTML]{000000} 0.79 \\
SleepLessThan7Hours & {\cellcolor[HTML]{F7FCF5}} \color[HTML]{000000} 0.39 & {\cellcolor[HTML]{EEF8EA}} \color[HTML]{000000} 0.46 & {\cellcolor[HTML]{E8F6E3}} \color[HTML]{000000} 0.51 & {\cellcolor[HTML]{E8F6E3}} \color[HTML]{000000} 0.51 & {\cellcolor[HTML]{BCE4B5}} \color[HTML]{000000} 0.71 & {\cellcolor[HTML]{C9EAC2}} \color[HTML]{000000} 0.66 & {\cellcolor[HTML]{B0DFAA}} \color[HTML]{000000} 0.75 & \bfseries {\cellcolor[HTML]{AEDEA7}} \color[HTML]{000000} 0.76 \\
Smoking & {\cellcolor[HTML]{F7FCF5}} \color[HTML]{000000} 0.44 & {\cellcolor[HTML]{F0F9EC}} \color[HTML]{000000} 0.50 & {\cellcolor[HTML]{E8F6E4}} \color[HTML]{000000} 0.56 & {\cellcolor[HTML]{E9F7E5}} \color[HTML]{000000} 0.55 & {\cellcolor[HTML]{C1E6BA}} \color[HTML]{000000} 0.75 & {\cellcolor[HTML]{C6E8BF}} \color[HTML]{000000} 0.73 & {\cellcolor[HTML]{B0DFAA}} \color[HTML]{000000} 0.81 & \bfseries {\cellcolor[HTML]{AEDEA7}} \color[HTML]{000000} 0.82 \\
Income & {\cellcolor[HTML]{F7FCF5}} \color[HTML]{000000} 0.34 & {\cellcolor[HTML]{EFF9EC}} \color[HTML]{000000} 0.40 & {\cellcolor[HTML]{EBF7E7}} \color[HTML]{000000} 0.43 & {\cellcolor[HTML]{ECF8E8}} \color[HTML]{000000} 0.42 & {\cellcolor[HTML]{B6E2AF}} \color[HTML]{000000} 0.66 & {\cellcolor[HTML]{BCE4B5}} \color[HTML]{000000} 0.64 & {\cellcolor[HTML]{B1E0AB}} \color[HTML]{000000} 0.68 & \bfseries {\cellcolor[HTML]{AEDEA7}} \color[HTML]{000000} 0.69 \\
HomeValue & {\cellcolor[HTML]{F7FCF5}} \color[HTML]{000000} 0.66 & {\cellcolor[HTML]{F2FAF0}} \color[HTML]{000000} 0.68 & {\cellcolor[HTML]{F7FCF5}} \color[HTML]{000000} 0.66 & {\cellcolor[HTML]{F7FCF5}} \color[HTML]{000000} 0.66 & {\cellcolor[HTML]{C3E7BC}} \color[HTML]{000000} 0.81 & {\cellcolor[HTML]{DAF0D4}} \color[HTML]{000000} 0.76 & {\cellcolor[HTML]{B4E1AD}} \color[HTML]{000000} 0.84 & \bfseries {\cellcolor[HTML]{AEDEA7}} \color[HTML]{000000} 0.85 \\
NightLights & {\cellcolor[HTML]{F5FBF2}} \color[HTML]{000000} 0.57 & {\cellcolor[HTML]{F7FCF5}} \color[HTML]{000000} 0.55 & {\cellcolor[HTML]{D3EECD}} \color[HTML]{000000} 0.78 & {\cellcolor[HTML]{CEECC8}} \color[HTML]{000000} 0.80 & {\cellcolor[HTML]{BBE4B4}} \color[HTML]{000000} 0.88 & {\cellcolor[HTML]{B4E1AD}} \color[HTML]{000000} 0.91 & \bfseries {\cellcolor[HTML]{AEDEA7}} \color[HTML]{000000} 0.93 & \bfseries {\cellcolor[HTML]{AEDEA7}} \color[HTML]{000000} 0.93 \\
PopulationDensity & {\cellcolor[HTML]{F4FBF1}} \color[HTML]{000000} 0.64 & {\cellcolor[HTML]{F7FCF5}} \color[HTML]{000000} 0.62 & {\cellcolor[HTML]{E1F3DC}} \color[HTML]{000000} 0.74 & {\cellcolor[HTML]{D6EFD0}} \color[HTML]{000000} 0.78 & {\cellcolor[HTML]{C3E7BC}} \color[HTML]{000000} 0.84 & {\cellcolor[HTML]{B5E1AE}} \color[HTML]{000000} 0.88 & \bfseries {\cellcolor[HTML]{AEDEA7}} \color[HTML]{000000} 0.90 & \bfseries {\cellcolor[HTML]{AEDEA7}} \color[HTML]{000000} 0.90 \\
TreeCover & {\cellcolor[HTML]{E5F5E0}} \color[HTML]{000000} 0.70 & {\cellcolor[HTML]{D2EDCC}} \color[HTML]{000000} 0.75 & {\cellcolor[HTML]{CEECC8}} \color[HTML]{000000} 0.76 & {\cellcolor[HTML]{DAF0D4}} \color[HTML]{000000} 0.73 & {\cellcolor[HTML]{F7FCF5}} \color[HTML]{000000} 0.62 & {\cellcolor[HTML]{F0F9ED}} \color[HTML]{000000} 0.65 & {\cellcolor[HTML]{C1E6BA}} \color[HTML]{000000} 0.79 & \bfseries {\cellcolor[HTML]{AEDEA7}} \color[HTML]{000000} 0.83 \\
Elevation & {\cellcolor[HTML]{BDE5B6}} \color[HTML]{000000} 0.94 & {\cellcolor[HTML]{B5E1AE}} \color[HTML]{000000} 0.96 & {\cellcolor[HTML]{C8E9C1}} \color[HTML]{000000} 0.91 & {\cellcolor[HTML]{B5E1AE}} \color[HTML]{000000} 0.96 & {\cellcolor[HTML]{EFF9EB}} \color[HTML]{000000} 0.76 & {\cellcolor[HTML]{F7FCF5}} \color[HTML]{000000} 0.71 & {\cellcolor[HTML]{B5E1AE}} \color[HTML]{000000} 0.96 & \bfseries {\cellcolor[HTML]{AEDEA7}} \color[HTML]{000000} 0.98 \\
All metrics (mean) & {\cellcolor[HTML]{F7FCF5}} \color[HTML]{000000} 0.43 & {\cellcolor[HTML]{F1FAEE}} \color[HTML]{000000} 0.47 & {\cellcolor[HTML]{EBF7E7}} \color[HTML]{000000} 0.52 & {\cellcolor[HTML]{EBF7E7}} \color[HTML]{000000} 0.52 & {\cellcolor[HTML]{CBEAC4}} \color[HTML]{000000} 0.67 & {\cellcolor[HTML]{CBEAC4}} \color[HTML]{000000} 0.67 & {\cellcolor[HTML]{B4E1AD}} \color[HTML]{000000} 0.75 & \bfseries {\cellcolor[HTML]{AEDEA7}} \color[HTML]{000000} 0.77 \\
\end{tabular}
\end{table}

\begin{table}
\fontsize{10pt}{10pt}\selectfont
\caption{\textbf{Extrapolation R2 with optimal bolded (Higher $R^2$ is Better).} We present our full set of experiments for extrapolation across all 27 downstream tasks including health, socioeconomic, and environment categories using the $R^2$ metric (a higher value shows the respective model better accounts for the variance in the groundtruth target variable labels). These experiments compare inverse distance weighted interpolation (IDW), SatCLIP embeddings, GeoCLIP embeddings, our PDFM embeddings and its subcomponents (Weather \& Air Quality, Trends, Maps, and Busyness), as well as a concatenation of PDFM and SatCLIP embeddings. This table indicates the model column with the optimal $R^2$ performance in bold and higher performance with darker green.}
\label{table:extrapolation_full}
\begin{tabular}{lrrrrrrrr}
 & IDW & SatCLIP & GeoCLIP & \makecell{Weather \\ \& AQ} & Trends & Maps & PDFM & \makecell{PDFM w/ \\ SatCLIP} \\
HighCholesterol & {\cellcolor[HTML]{EDF8E9}} \color[HTML]{000000} -0.01 & {\cellcolor[HTML]{F0F9EC}} \color[HTML]{000000} -0.05 & {\cellcolor[HTML]{DBF1D5}} \color[HTML]{000000} 0.17 & {\cellcolor[HTML]{F7FCF5}} \color[HTML]{000000} -0.15 & {\cellcolor[HTML]{CCEBC6}} \color[HTML]{000000} 0.28 & {\cellcolor[HTML]{B0DFAA}} \color[HTML]{000000} 0.47 & {\cellcolor[HTML]{B0DFAA}} \color[HTML]{000000} 0.47 & \bfseries {\cellcolor[HTML]{AEDEA7}} \color[HTML]{000000} 0.48 \\
PhysicalHealthNotGood & {\cellcolor[HTML]{F7FCF5}} \color[HTML]{000000} -0.22 & {\cellcolor[HTML]{E4F5DF}} \color[HTML]{000000} 0.15 & {\cellcolor[HTML]{C9EAC2}} \color[HTML]{000000} 0.46 & {\cellcolor[HTML]{D7EFD1}} \color[HTML]{000000} 0.30 & {\cellcolor[HTML]{B7E2B1}} \color[HTML]{000000} 0.63 & {\cellcolor[HTML]{B7E2B1}} \color[HTML]{000000} 0.62 & {\cellcolor[HTML]{AFDFA8}} \color[HTML]{000000} 0.70 & \bfseries {\cellcolor[HTML]{AEDEA7}} \color[HTML]{000000} 0.71 \\
Stroke & {\cellcolor[HTML]{F7FCF5}} \color[HTML]{000000} -0.30 & {\cellcolor[HTML]{E8F6E3}} \color[HTML]{000000} -0.01 & {\cellcolor[HTML]{CBEBC5}} \color[HTML]{000000} 0.33 & {\cellcolor[HTML]{D6EFD0}} \color[HTML]{000000} 0.22 & {\cellcolor[HTML]{BAE3B3}} \color[HTML]{000000} 0.50 & {\cellcolor[HTML]{BAE3B3}} \color[HTML]{000000} 0.51 & {\cellcolor[HTML]{B0DFAA}} \color[HTML]{000000} 0.59 & \bfseries {\cellcolor[HTML]{AEDEA7}} \color[HTML]{000000} 0.61 \\
BingeDrinking & {\cellcolor[HTML]{E9F7E5}} \color[HTML]{000000} -0.06 & {\cellcolor[HTML]{F7FCF5}} \color[HTML]{000000} -0.21 & {\cellcolor[HTML]{D0EDCA}} \color[HTML]{000000} 0.12 & {\cellcolor[HTML]{E8F6E3}} \color[HTML]{000000} -0.04 & {\cellcolor[HTML]{BDE5B6}} \color[HTML]{000000} 0.23 & {\cellcolor[HTML]{C2E7BB}} \color[HTML]{000000} 0.21 & \bfseries {\cellcolor[HTML]{AEDEA7}} \color[HTML]{000000} 0.31 & {\cellcolor[HTML]{BDE5B6}} \color[HTML]{000000} 0.23 \\
PhysicalInactivity & {\cellcolor[HTML]{F7FCF5}} \color[HTML]{000000} -0.38 & {\cellcolor[HTML]{DBF1D5}} \color[HTML]{000000} 0.13 & {\cellcolor[HTML]{C7E9C0}} \color[HTML]{000000} 0.37 & {\cellcolor[HTML]{D2EDCC}} \color[HTML]{000000} 0.23 & {\cellcolor[HTML]{B4E1AD}} \color[HTML]{000000} 0.56 & \bfseries {\cellcolor[HTML]{AEDEA7}} \color[HTML]{000000} 0.62 & {\cellcolor[HTML]{B2E0AC}} \color[HTML]{000000} 0.58 & {\cellcolor[HTML]{B0DFAA}} \color[HTML]{000000} 0.60 \\
ReceivedAnnualCheckup & {\cellcolor[HTML]{CFECC9}} \color[HTML]{000000} 0.51 & {\cellcolor[HTML]{E9F7E5}} \color[HTML]{000000} 0.43 & {\cellcolor[HTML]{F7FCF5}} \color[HTML]{000000} 0.36 & {\cellcolor[HTML]{C4E8BD}} \color[HTML]{000000} 0.54 & {\cellcolor[HTML]{D6EFD0}} \color[HTML]{000000} 0.49 & {\cellcolor[HTML]{F7FCF5}} \color[HTML]{000000} 0.36 & \bfseries {\cellcolor[HTML]{AEDEA7}} \color[HTML]{000000} 0.59 & {\cellcolor[HTML]{C4E8BD}} \color[HTML]{000000} 0.54 \\
Cancer & {\cellcolor[HTML]{F7FCF5}} \color[HTML]{000000} -0.16 & {\cellcolor[HTML]{F5FBF3}} \color[HTML]{000000} -0.13 & {\cellcolor[HTML]{D6EFD0}} \color[HTML]{000000} 0.27 & {\cellcolor[HTML]{E0F3DB}} \color[HTML]{000000} 0.17 & {\cellcolor[HTML]{BBE4B4}} \color[HTML]{000000} 0.50 & {\cellcolor[HTML]{B6E2AF}} \color[HTML]{000000} 0.54 & \bfseries {\cellcolor[HTML]{AEDEA7}} \color[HTML]{000000} 0.60 & {\cellcolor[HTML]{AFDFA8}} \color[HTML]{000000} 0.59 \\
Diabetes & {\cellcolor[HTML]{F7FCF5}} \color[HTML]{000000} -0.04 & {\cellcolor[HTML]{EEF8EA}} \color[HTML]{000000} 0.09 & {\cellcolor[HTML]{D7EFD1}} \color[HTML]{000000} 0.34 & {\cellcolor[HTML]{E1F3DC}} \color[HTML]{000000} 0.25 & {\cellcolor[HTML]{BEE5B8}} \color[HTML]{000000} 0.53 & {\cellcolor[HTML]{BBE4B4}} \color[HTML]{000000} 0.55 & \bfseries {\cellcolor[HTML]{AEDEA7}} \color[HTML]{000000} 0.64 & {\cellcolor[HTML]{B1E0AB}} \color[HTML]{000000} 0.62 \\
MentalHealthNotGood & {\cellcolor[HTML]{F7FCF5}} \color[HTML]{000000} -0.02 & {\cellcolor[HTML]{E9F7E5}} \color[HTML]{000000} 0.18 & {\cellcolor[HTML]{D7EFD1}} \color[HTML]{000000} 0.35 & {\cellcolor[HTML]{E9F7E5}} \color[HTML]{000000} 0.18 & {\cellcolor[HTML]{C0E6B9}} \color[HTML]{000000} 0.53 & {\cellcolor[HTML]{B6E2AF}} \color[HTML]{000000} 0.60 & {\cellcolor[HTML]{AFDFA8}} \color[HTML]{000000} 0.64 & \bfseries {\cellcolor[HTML]{AEDEA7}} \color[HTML]{000000} 0.65 \\
CoronaryHeartDisease & {\cellcolor[HTML]{F7FCF5}} \color[HTML]{000000} -0.41 & {\cellcolor[HTML]{E7F6E2}} \color[HTML]{000000} -0.06 & {\cellcolor[HTML]{C7E9C0}} \color[HTML]{000000} 0.35 & {\cellcolor[HTML]{D2EDCC}} \color[HTML]{000000} 0.21 & {\cellcolor[HTML]{B8E3B2}} \color[HTML]{000000} 0.49 & {\cellcolor[HTML]{B7E2B1}} \color[HTML]{000000} 0.51 & {\cellcolor[HTML]{B1E0AB}} \color[HTML]{000000} 0.57 & \bfseries {\cellcolor[HTML]{AEDEA7}} \color[HTML]{000000} 0.60 \\
HighBloodPressure & {\cellcolor[HTML]{F7FCF5}} \color[HTML]{000000} 0.03 & {\cellcolor[HTML]{F3FAF0}} \color[HTML]{000000} 0.09 & {\cellcolor[HTML]{D7EFD1}} \color[HTML]{000000} 0.40 & {\cellcolor[HTML]{E5F5E0}} \color[HTML]{000000} 0.28 & {\cellcolor[HTML]{C3E7BC}} \color[HTML]{000000} 0.55 & {\cellcolor[HTML]{BBE4B4}} \color[HTML]{000000} 0.61 & {\cellcolor[HTML]{B2E0AC}} \color[HTML]{000000} 0.66 & \bfseries {\cellcolor[HTML]{AEDEA7}} \color[HTML]{000000} 0.69 \\
CholesterolScreening & {\cellcolor[HTML]{F7FCF5}} \color[HTML]{000000} -0.17 & {\cellcolor[HTML]{EBF7E7}} \color[HTML]{000000} -0.07 & {\cellcolor[HTML]{DAF0D4}} \color[HTML]{000000} 0.04 & {\cellcolor[HTML]{EDF8EA}} \color[HTML]{000000} -0.09 & {\cellcolor[HTML]{BDE5B6}} \color[HTML]{000000} 0.17 & \bfseries {\cellcolor[HTML]{AEDEA7}} \color[HTML]{000000} 0.23 & \bfseries {\cellcolor[HTML]{AEDEA7}} \color[HTML]{000000} 0.23 & {\cellcolor[HTML]{BDE5B6}} \color[HTML]{000000} 0.17 \\
DentalVisit & {\cellcolor[HTML]{F7FCF5}} \color[HTML]{000000} -0.19 & {\cellcolor[HTML]{E0F3DB}} \color[HTML]{000000} 0.19 & {\cellcolor[HTML]{CDECC7}} \color[HTML]{000000} 0.40 & {\cellcolor[HTML]{D9F0D3}} \color[HTML]{000000} 0.27 & {\cellcolor[HTML]{B2E0AC}} \color[HTML]{000000} 0.64 & {\cellcolor[HTML]{B4E1AD}} \color[HTML]{000000} 0.63 & \bfseries {\cellcolor[HTML]{AEDEA7}} \color[HTML]{000000} 0.68 & {\cellcolor[HTML]{B1E0AB}} \color[HTML]{000000} 0.65 \\
Asthma & {\cellcolor[HTML]{F7FCF5}} \color[HTML]{000000} -0.30 & {\cellcolor[HTML]{E5F5E1}} \color[HTML]{000000} -0.02 & {\cellcolor[HTML]{C3E7BC}} \color[HTML]{000000} 0.30 & {\cellcolor[HTML]{D2EDCC}} \color[HTML]{000000} 0.17 & {\cellcolor[HTML]{B4E1AD}} \color[HTML]{000000} 0.42 & {\cellcolor[HTML]{B7E2B1}} \color[HTML]{000000} 0.39 & \bfseries {\cellcolor[HTML]{AEDEA7}} \color[HTML]{000000} 0.46 & {\cellcolor[HTML]{B0DFAA}} \color[HTML]{000000} 0.44 \\
ChronicKidneyDisease & {\cellcolor[HTML]{F7FCF5}} \color[HTML]{000000} -0.42 & {\cellcolor[HTML]{E3F4DE}} \color[HTML]{000000} -0.02 & {\cellcolor[HTML]{CBEAC4}} \color[HTML]{000000} 0.28 & {\cellcolor[HTML]{D1EDCB}} \color[HTML]{000000} 0.20 & {\cellcolor[HTML]{BBE4B4}} \color[HTML]{000000} 0.44 & {\cellcolor[HTML]{B8E3B2}} \color[HTML]{000000} 0.47 & {\cellcolor[HTML]{AFDFA8}} \color[HTML]{000000} 0.56 & \bfseries {\cellcolor[HTML]{AEDEA7}} \color[HTML]{000000} 0.57 \\
Arthritis & {\cellcolor[HTML]{F7FCF5}} \color[HTML]{000000} -0.09 & {\cellcolor[HTML]{EFF9EB}} \color[HTML]{000000} 0.05 & {\cellcolor[HTML]{CCEBC6}} \color[HTML]{000000} 0.45 & {\cellcolor[HTML]{D3EECD}} \color[HTML]{000000} 0.39 & {\cellcolor[HTML]{BCE4B5}} \color[HTML]{000000} 0.59 & {\cellcolor[HTML]{B6E2AF}} \color[HTML]{000000} 0.64 & {\cellcolor[HTML]{AFDFA8}} \color[HTML]{000000} 0.69 & \bfseries {\cellcolor[HTML]{AEDEA7}} \color[HTML]{000000} 0.70 \\
COPD & {\cellcolor[HTML]{F7FCF5}} \color[HTML]{000000} -0.18 & {\cellcolor[HTML]{E4F5DF}} \color[HTML]{000000} 0.16 & {\cellcolor[HTML]{C7E9C0}} \color[HTML]{000000} 0.47 & {\cellcolor[HTML]{D1EDCB}} \color[HTML]{000000} 0.36 & {\cellcolor[HTML]{B7E2B1}} \color[HTML]{000000} 0.60 & {\cellcolor[HTML]{B7E2B1}} \color[HTML]{000000} 0.60 & {\cellcolor[HTML]{B1E0AB}} \color[HTML]{000000} 0.65 & \bfseries {\cellcolor[HTML]{AEDEA7}} \color[HTML]{000000} 0.68 \\
BPMed & {\cellcolor[HTML]{F7FCF5}} \color[HTML]{000000} -0.08 & {\cellcolor[HTML]{EFF9EC}} \color[HTML]{000000} 0.06 & {\cellcolor[HTML]{E6F5E1}} \color[HTML]{000000} 0.21 & {\cellcolor[HTML]{EEF8EA}} \color[HTML]{000000} 0.08 & {\cellcolor[HTML]{D8F0D2}} \color[HTML]{000000} 0.36 & {\cellcolor[HTML]{B4E1AD}} \color[HTML]{000000} 0.68 & \bfseries {\cellcolor[HTML]{AEDEA7}} \color[HTML]{000000} 0.73 & {\cellcolor[HTML]{AFDFA8}} \color[HTML]{000000} 0.72 \\
Obesity & {\cellcolor[HTML]{F7FCF5}} \color[HTML]{000000} -0.21 & {\cellcolor[HTML]{E8F6E4}} \color[HTML]{000000} 0.05 & {\cellcolor[HTML]{C9EAC2}} \color[HTML]{000000} 0.41 & {\cellcolor[HTML]{DEF2D9}} \color[HTML]{000000} 0.18 & {\cellcolor[HTML]{AFDFA8}} \color[HTML]{000000} 0.62 & \bfseries {\cellcolor[HTML]{AEDEA7}} \color[HTML]{000000} 0.63 & \bfseries {\cellcolor[HTML]{AEDEA7}} \color[HTML]{000000} 0.63 & \bfseries {\cellcolor[HTML]{AEDEA7}} \color[HTML]{000000} 0.63 \\
SleepLessThan7Hours & {\cellcolor[HTML]{F7FCF5}} \color[HTML]{000000} -0.10 & {\cellcolor[HTML]{E6F5E1}} \color[HTML]{000000} 0.14 & {\cellcolor[HTML]{D3EECD}} \color[HTML]{000000} 0.31 & {\cellcolor[HTML]{E5F5E1}} \color[HTML]{000000} 0.15 & {\cellcolor[HTML]{B1E0AB}} \color[HTML]{000000} 0.56 & {\cellcolor[HTML]{BEE5B8}} \color[HTML]{000000} 0.47 & \bfseries {\cellcolor[HTML]{AEDEA7}} \color[HTML]{000000} 0.58 & {\cellcolor[HTML]{B2E0AC}} \color[HTML]{000000} 0.55 \\
Smoking & {\cellcolor[HTML]{F7FCF5}} \color[HTML]{000000} 0.02 & {\cellcolor[HTML]{E9F7E5}} \color[HTML]{000000} 0.22 & {\cellcolor[HTML]{CBEAC4}} \color[HTML]{000000} 0.49 & {\cellcolor[HTML]{D4EECE}} \color[HTML]{000000} 0.41 & {\cellcolor[HTML]{B4E1AD}} \color[HTML]{000000} 0.64 & {\cellcolor[HTML]{B6E2AF}} \color[HTML]{000000} 0.63 & {\cellcolor[HTML]{B1E0AB}} \color[HTML]{000000} 0.66 & \bfseries {\cellcolor[HTML]{AEDEA7}} \color[HTML]{000000} 0.68 \\
Income & {\cellcolor[HTML]{F7FCF5}} \color[HTML]{000000} -0.09 & {\cellcolor[HTML]{E9F7E5}} \color[HTML]{000000} 0.11 & {\cellcolor[HTML]{CFECC9}} \color[HTML]{000000} 0.35 & {\cellcolor[HTML]{E5F5E0}} \color[HTML]{000000} 0.17 & \bfseries {\cellcolor[HTML]{AEDEA7}} \color[HTML]{000000} 0.59 & {\cellcolor[HTML]{B1E0AB}} \color[HTML]{000000} 0.57 & \bfseries {\cellcolor[HTML]{AEDEA7}} \color[HTML]{000000} 0.59 & \bfseries {\cellcolor[HTML]{AEDEA7}} \color[HTML]{000000} 0.59 \\
HomeValue & {\cellcolor[HTML]{F7FCF5}} \color[HTML]{000000} -0.09 & {\cellcolor[HTML]{E9F7E5}} \color[HTML]{000000} 0.10 & {\cellcolor[HTML]{D3EECD}} \color[HTML]{000000} 0.29 & {\cellcolor[HTML]{E0F3DB}} \color[HTML]{000000} 0.19 & {\cellcolor[HTML]{B1E0AB}} \color[HTML]{000000} 0.52 & {\cellcolor[HTML]{B1E0AB}} \color[HTML]{000000} 0.52 & \bfseries {\cellcolor[HTML]{AEDEA7}} \color[HTML]{000000} 0.54 & {\cellcolor[HTML]{BCE4B5}} \color[HTML]{000000} 0.45 \\
NightLights & {\cellcolor[HTML]{F7FCF5}} \color[HTML]{000000} 0.11 & {\cellcolor[HTML]{EFF9EC}} \color[HTML]{000000} 0.25 & {\cellcolor[HTML]{BCE4B5}} \color[HTML]{000000} 0.80 & {\cellcolor[HTML]{CBEAC4}} \color[HTML]{000000} 0.68 & {\cellcolor[HTML]{B6E2AF}} \color[HTML]{000000} 0.85 & {\cellcolor[HTML]{B0DFAA}} \color[HTML]{000000} 0.89 & \bfseries {\cellcolor[HTML]{AEDEA7}} \color[HTML]{000000} 0.91 & \bfseries {\cellcolor[HTML]{AEDEA7}} \color[HTML]{000000} 0.91 \\
PopulationDensity & {\cellcolor[HTML]{F7FCF5}} \color[HTML]{000000} 0.20 & {\cellcolor[HTML]{F6FCF4}} \color[HTML]{000000} 0.22 & {\cellcolor[HTML]{C3E7BC}} \color[HTML]{000000} 0.74 & {\cellcolor[HTML]{CDECC7}} \color[HTML]{000000} 0.66 & {\cellcolor[HTML]{BBE4B4}} \color[HTML]{000000} 0.79 & {\cellcolor[HTML]{AFDFA8}} \color[HTML]{000000} 0.87 & \bfseries {\cellcolor[HTML]{AEDEA7}} \color[HTML]{000000} 0.88 & \bfseries {\cellcolor[HTML]{AEDEA7}} \color[HTML]{000000} 0.88 \\
TreeCover & {\cellcolor[HTML]{F7FCF5}} \color[HTML]{000000} 0.36 & {\cellcolor[HTML]{E7F6E3}} \color[HTML]{000000} 0.47 & \bfseries {\cellcolor[HTML]{AEDEA7}} \color[HTML]{000000} 0.69 & {\cellcolor[HTML]{DCF2D7}} \color[HTML]{000000} 0.52 & {\cellcolor[HTML]{EAF7E6}} \color[HTML]{000000} 0.45 & {\cellcolor[HTML]{DCF2D7}} \color[HTML]{000000} 0.52 & {\cellcolor[HTML]{C7E9C0}} \color[HTML]{000000} 0.61 & {\cellcolor[HTML]{B7E2B1}} \color[HTML]{000000} 0.66 \\
Elevation & {\cellcolor[HTML]{F7FCF5}} \color[HTML]{000000} 0.41 & {\cellcolor[HTML]{CBEBC5}} \color[HTML]{000000} 0.69 & \bfseries {\cellcolor[HTML]{AEDEA7}} \color[HTML]{000000} 0.81 & {\cellcolor[HTML]{B2E0AC}} \color[HTML]{000000} 0.79 & {\cellcolor[HTML]{E9F7E5}} \color[HTML]{000000} 0.53 & {\cellcolor[HTML]{EAF7E6}} \color[HTML]{000000} 0.52 & \bfseries {\cellcolor[HTML]{AEDEA7}} \color[HTML]{000000} 0.81 & {\cellcolor[HTML]{C0E6B9}} \color[HTML]{000000} 0.74 \\
All metrics (mean) & {\cellcolor[HTML]{F7FCF5}} \color[HTML]{000000} -0.07 & {\cellcolor[HTML]{EAF7E6}} \color[HTML]{000000} 0.12 & {\cellcolor[HTML]{CDECC7}} \color[HTML]{000000} 0.39 & {\cellcolor[HTML]{DBF1D6}} \color[HTML]{000000} 0.27 & {\cellcolor[HTML]{BBE4B4}} \color[HTML]{000000} 0.52 & {\cellcolor[HTML]{B7E2B1}} \color[HTML]{000000} 0.55 & \bfseries {\cellcolor[HTML]{AEDEA7}} \color[HTML]{000000} 0.61 & {\cellcolor[HTML]{AFDFA8}} \color[HTML]{000000} 0.60 \\
\end{tabular}
\end{table}

\begin{table}
\fontsize{10pt}{10pt}\selectfont
\caption{\textbf{Intracounty Super-Resolution Pearson's correlation coefficient ($r$) with optimal bolded (Higher $r$ is Better).} We present our full set of experiments for super-resolution across all 27 tasks using the Pearson's r metric (a higher value shows the respective model better accounts for the variance in the groundtruth target variable labels). These experiments compare inverse distance weighted interpolation (IDW), SatCLIP embeddings, GeoCLIP embeddings, our PDFM embeddings and its subcomponents  (Weather \& Air Quality, Trends, Maps, and Busyness), as well as a concatenation of PDFM and SatCLIP embeddings. This table indicates the model column with the optimal $r$ performance in bold and higher performance with darker green.}
\label{table:superresolution_full}
\begin{tabular}{lrrrrrrrr}
 & IDW & SatCLIP & GeoCLIP & \makecell{Weather \\ \& AQ} & Trends & Maps & PDFM & \makecell{PDFM w/ \\ SatCLIP} \\
HighCholesterol & {\cellcolor[HTML]{F4FBF1}} \color[HTML]{000000} 0.13 & {\cellcolor[HTML]{F7FCF5}} \color[HTML]{000000} 0.11 & {\cellcolor[HTML]{F0F9ED}} \color[HTML]{000000} 0.15 & {\cellcolor[HTML]{F5FBF3}} \color[HTML]{000000} 0.12 & {\cellcolor[HTML]{BAE3B3}} \color[HTML]{000000} 0.35 & {\cellcolor[HTML]{D7EFD1}} \color[HTML]{000000} 0.26 & {\cellcolor[HTML]{B1E0AB}} \color[HTML]{000000} 0.37 & \bfseries {\cellcolor[HTML]{AEDEA7}} \color[HTML]{000000} 0.38 \\
PhysicalHealthNotGood & {\cellcolor[HTML]{F6FCF4}} \color[HTML]{000000} 0.08 & {\cellcolor[HTML]{F1FAEE}} \color[HTML]{000000} 0.13 & {\cellcolor[HTML]{ECF8E8}} \color[HTML]{000000} 0.18 & {\cellcolor[HTML]{F7FCF5}} \color[HTML]{000000} 0.07 & \bfseries {\cellcolor[HTML]{AEDEA7}} \color[HTML]{000000} 0.55 & {\cellcolor[HTML]{CCEBC6}} \color[HTML]{000000} 0.40 & {\cellcolor[HTML]{B2E0AC}} \color[HTML]{000000} 0.53 & {\cellcolor[HTML]{B0DFAA}} \color[HTML]{000000} 0.54 \\
Stroke & {\cellcolor[HTML]{EDF8EA}} \color[HTML]{000000} 0.14 & {\cellcolor[HTML]{F2FAF0}} \color[HTML]{000000} 0.10 & {\cellcolor[HTML]{ECF8E8}} \color[HTML]{000000} 0.15 & {\cellcolor[HTML]{F7FCF5}} \color[HTML]{000000} 0.06 & \bfseries {\cellcolor[HTML]{AEDEA7}} \color[HTML]{000000} 0.46 & {\cellcolor[HTML]{CFECC9}} \color[HTML]{000000} 0.32 & \bfseries {\cellcolor[HTML]{AEDEA7}} \color[HTML]{000000} 0.46 & \bfseries {\cellcolor[HTML]{AEDEA7}} \color[HTML]{000000} 0.46 \\
BingeDrinking & {\cellcolor[HTML]{EDF8EA}} \color[HTML]{000000} 0.07 & {\cellcolor[HTML]{ECF8E8}} \color[HTML]{000000} 0.08 & {\cellcolor[HTML]{EFF9EC}} \color[HTML]{000000} 0.06 & {\cellcolor[HTML]{F7FCF5}} \color[HTML]{000000} 0.01 & {\cellcolor[HTML]{B8E3B2}} \color[HTML]{000000} 0.28 & {\cellcolor[HTML]{CEECC8}} \color[HTML]{000000} 0.21 & {\cellcolor[HTML]{B5E1AE}} \color[HTML]{000000} 0.29 & \bfseries {\cellcolor[HTML]{AEDEA7}} \color[HTML]{000000} 0.31 \\
PhysicalInactivity & {\cellcolor[HTML]{F7FCF5}} \color[HTML]{000000} 0.03 & {\cellcolor[HTML]{F3FAF0}} \color[HTML]{000000} 0.08 & {\cellcolor[HTML]{E8F6E4}} \color[HTML]{000000} 0.20 & {\cellcolor[HTML]{F6FCF4}} \color[HTML]{000000} 0.04 & {\cellcolor[HTML]{B0DFAA}} \color[HTML]{000000} 0.57 & {\cellcolor[HTML]{C9EAC2}} \color[HTML]{000000} 0.43 & {\cellcolor[HTML]{B1E0AB}} \color[HTML]{000000} 0.56 & \bfseries {\cellcolor[HTML]{AEDEA7}} \color[HTML]{000000} 0.58 \\
ReceivedAnnualCheckup & {\cellcolor[HTML]{ECF8E8}} \color[HTML]{000000} 0.11 & {\cellcolor[HTML]{F1FAEE}} \color[HTML]{000000} 0.08 & {\cellcolor[HTML]{F7FCF5}} \color[HTML]{000000} 0.04 & {\cellcolor[HTML]{F4FBF2}} \color[HTML]{000000} 0.06 & {\cellcolor[HTML]{B8E3B2}} \color[HTML]{000000} 0.31 & {\cellcolor[HTML]{D9F0D3}} \color[HTML]{000000} 0.20 & {\cellcolor[HTML]{B1E0AB}} \color[HTML]{000000} 0.33 & \bfseries {\cellcolor[HTML]{AEDEA7}} \color[HTML]{000000} 0.34 \\
Cancer & {\cellcolor[HTML]{F7FCF5}} \color[HTML]{000000} 0.16 & {\cellcolor[HTML]{F6FCF4}} \color[HTML]{000000} 0.17 & {\cellcolor[HTML]{F4FBF2}} \color[HTML]{000000} 0.18 & {\cellcolor[HTML]{F6FCF4}} \color[HTML]{000000} 0.17 & {\cellcolor[HTML]{B7E2B1}} \color[HTML]{000000} 0.44 & {\cellcolor[HTML]{D2EDCC}} \color[HTML]{000000} 0.35 & \bfseries {\cellcolor[HTML]{AEDEA7}} \color[HTML]{000000} 0.47 & \bfseries {\cellcolor[HTML]{AEDEA7}} \color[HTML]{000000} 0.47 \\
Diabetes & {\cellcolor[HTML]{F3FAF0}} \color[HTML]{000000} 0.11 & {\cellcolor[HTML]{F3FAF0}} \color[HTML]{000000} 0.11 & {\cellcolor[HTML]{F0F9EC}} \color[HTML]{000000} 0.14 & {\cellcolor[HTML]{F7FCF5}} \color[HTML]{000000} 0.07 & \bfseries {\cellcolor[HTML]{AEDEA7}} \color[HTML]{000000} 0.50 & {\cellcolor[HTML]{CBEBC5}} \color[HTML]{000000} 0.37 & \bfseries {\cellcolor[HTML]{AEDEA7}} \color[HTML]{000000} 0.50 & \bfseries {\cellcolor[HTML]{AEDEA7}} \color[HTML]{000000} 0.50 \\
MentalHealthNotGood & {\cellcolor[HTML]{F7FCF5}} \color[HTML]{000000} 0.06 & {\cellcolor[HTML]{F4FBF1}} \color[HTML]{000000} 0.10 & {\cellcolor[HTML]{EDF8EA}} \color[HTML]{000000} 0.16 & {\cellcolor[HTML]{F5FBF3}} \color[HTML]{000000} 0.08 & {\cellcolor[HTML]{B0DFAA}} \color[HTML]{000000} 0.54 & {\cellcolor[HTML]{CDECC7}} \color[HTML]{000000} 0.39 & {\cellcolor[HTML]{B0DFAA}} \color[HTML]{000000} 0.54 & \bfseries {\cellcolor[HTML]{AEDEA7}} \color[HTML]{000000} 0.55 \\
CoronaryHeartDisease & {\cellcolor[HTML]{F0F9ED}} \color[HTML]{000000} 0.13 & {\cellcolor[HTML]{F6FCF4}} \color[HTML]{000000} 0.09 & {\cellcolor[HTML]{EFF9EB}} \color[HTML]{000000} 0.14 & {\cellcolor[HTML]{F7FCF5}} \color[HTML]{000000} 0.08 & {\cellcolor[HTML]{B1E0AB}} \color[HTML]{000000} 0.40 & {\cellcolor[HTML]{D5EFCF}} \color[HTML]{000000} 0.27 & {\cellcolor[HTML]{B1E0AB}} \color[HTML]{000000} 0.40 & \bfseries {\cellcolor[HTML]{AEDEA7}} \color[HTML]{000000} 0.41 \\
HighBloodPressure & {\cellcolor[HTML]{EDF8EA}} \color[HTML]{000000} 0.16 & {\cellcolor[HTML]{F5FBF2}} \color[HTML]{000000} 0.11 & {\cellcolor[HTML]{F2FAEF}} \color[HTML]{000000} 0.13 & {\cellcolor[HTML]{F7FCF5}} \color[HTML]{000000} 0.09 & {\cellcolor[HTML]{B6E2AF}} \color[HTML]{000000} 0.41 & {\cellcolor[HTML]{D1EDCB}} \color[HTML]{000000} 0.31 & \bfseries {\cellcolor[HTML]{AEDEA7}} \color[HTML]{000000} 0.44 & \bfseries {\cellcolor[HTML]{AEDEA7}} \color[HTML]{000000} 0.44 \\
CholesterolScreening & {\cellcolor[HTML]{F4FBF1}} \color[HTML]{000000} 0.07 & {\cellcolor[HTML]{F5FBF2}} \color[HTML]{000000} 0.06 & {\cellcolor[HTML]{EFF9EC}} \color[HTML]{000000} 0.11 & {\cellcolor[HTML]{F7FCF5}} \color[HTML]{000000} 0.04 & {\cellcolor[HTML]{BBE4B4}} \color[HTML]{000000} 0.39 & {\cellcolor[HTML]{C7E9C0}} \color[HTML]{000000} 0.34 & {\cellcolor[HTML]{BBE4B4}} \color[HTML]{000000} 0.39 & \bfseries {\cellcolor[HTML]{AEDEA7}} \color[HTML]{000000} 0.44 \\
DentalVisit & {\cellcolor[HTML]{F7FCF5}} \color[HTML]{000000} 0.02 & {\cellcolor[HTML]{F1FAEE}} \color[HTML]{000000} 0.10 & {\cellcolor[HTML]{ECF8E8}} \color[HTML]{000000} 0.16 & {\cellcolor[HTML]{F2FAF0}} \color[HTML]{000000} 0.08 & {\cellcolor[HTML]{B1E0AB}} \color[HTML]{000000} 0.59 & {\cellcolor[HTML]{C3E7BC}} \color[HTML]{000000} 0.49 & {\cellcolor[HTML]{B0DFAA}} \color[HTML]{000000} 0.60 & \bfseries {\cellcolor[HTML]{AEDEA7}} \color[HTML]{000000} 0.61 \\
Asthma & {\cellcolor[HTML]{F1FAEE}} \color[HTML]{000000} 0.09 & {\cellcolor[HTML]{EDF8EA}} \color[HTML]{000000} 0.12 & {\cellcolor[HTML]{E7F6E3}} \color[HTML]{000000} 0.18 & {\cellcolor[HTML]{F7FCF5}} \color[HTML]{000000} 0.03 & {\cellcolor[HTML]{B0DFAA}} \color[HTML]{000000} 0.47 & {\cellcolor[HTML]{C9EAC2}} \color[HTML]{000000} 0.36 & \bfseries {\cellcolor[HTML]{AEDEA7}} \color[HTML]{000000} 0.48 & \bfseries {\cellcolor[HTML]{AEDEA7}} \color[HTML]{000000} 0.48 \\
ChronicKidneyDisease & {\cellcolor[HTML]{ECF8E8}} \color[HTML]{000000} 0.15 & {\cellcolor[HTML]{F2FAEF}} \color[HTML]{000000} 0.10 & {\cellcolor[HTML]{ECF8E8}} \color[HTML]{000000} 0.15 & {\cellcolor[HTML]{F7FCF5}} \color[HTML]{000000} 0.06 & {\cellcolor[HTML]{B0DFAA}} \color[HTML]{000000} 0.42 & {\cellcolor[HTML]{CFECC9}} \color[HTML]{000000} 0.30 & \bfseries {\cellcolor[HTML]{AEDEA7}} \color[HTML]{000000} 0.43 & \bfseries {\cellcolor[HTML]{AEDEA7}} \color[HTML]{000000} 0.43 \\
Arthritis & {\cellcolor[HTML]{F0F9EC}} \color[HTML]{000000} 0.18 & {\cellcolor[HTML]{F6FCF4}} \color[HTML]{000000} 0.14 & {\cellcolor[HTML]{F0F9EC}} \color[HTML]{000000} 0.18 & {\cellcolor[HTML]{F7FCF5}} \color[HTML]{000000} 0.13 & {\cellcolor[HTML]{B7E2B1}} \color[HTML]{000000} 0.41 & {\cellcolor[HTML]{DAF0D4}} \color[HTML]{000000} 0.29 & {\cellcolor[HTML]{B1E0AB}} \color[HTML]{000000} 0.43 & \bfseries {\cellcolor[HTML]{AEDEA7}} \color[HTML]{000000} 0.44 \\
COPD & {\cellcolor[HTML]{F1FAEE}} \color[HTML]{000000} 0.12 & {\cellcolor[HTML]{F0F9ED}} \color[HTML]{000000} 0.13 & {\cellcolor[HTML]{EBF7E7}} \color[HTML]{000000} 0.18 & {\cellcolor[HTML]{F7FCF5}} \color[HTML]{000000} 0.07 & {\cellcolor[HTML]{B2E0AC}} \color[HTML]{000000} 0.46 & {\cellcolor[HTML]{D0EDCA}} \color[HTML]{000000} 0.33 & {\cellcolor[HTML]{B2E0AC}} \color[HTML]{000000} 0.46 & \bfseries {\cellcolor[HTML]{AEDEA7}} \color[HTML]{000000} 0.48 \\
BPMed & {\cellcolor[HTML]{EFF9EB}} \color[HTML]{000000} 0.13 & {\cellcolor[HTML]{F5FBF3}} \color[HTML]{000000} 0.09 & {\cellcolor[HTML]{F0F9ED}} \color[HTML]{000000} 0.12 & {\cellcolor[HTML]{F7FCF5}} \color[HTML]{000000} 0.08 & {\cellcolor[HTML]{BDE5B6}} \color[HTML]{000000} 0.31 & {\cellcolor[HTML]{DAF0D4}} \color[HTML]{000000} 0.22 & {\cellcolor[HTML]{B1E0AB}} \color[HTML]{000000} 0.34 & \bfseries {\cellcolor[HTML]{AEDEA7}} \color[HTML]{000000} 0.35 \\
Obesity & {\cellcolor[HTML]{F7FCF5}} \color[HTML]{000000} 0.06 & {\cellcolor[HTML]{F0F9ED}} \color[HTML]{000000} 0.13 & {\cellcolor[HTML]{EFF9EC}} \color[HTML]{000000} 0.14 & {\cellcolor[HTML]{F4FBF2}} \color[HTML]{000000} 0.09 & \bfseries {\cellcolor[HTML]{AEDEA7}} \color[HTML]{000000} 0.54 & {\cellcolor[HTML]{D1EDCB}} \color[HTML]{000000} 0.36 & {\cellcolor[HTML]{B4E1AD}} \color[HTML]{000000} 0.51 & {\cellcolor[HTML]{B4E1AD}} \color[HTML]{000000} 0.51 \\
SleepLessThan7Hours & {\cellcolor[HTML]{F7FCF5}} \color[HTML]{000000} 0.07 & {\cellcolor[HTML]{F0F9ED}} \color[HTML]{000000} 0.14 & {\cellcolor[HTML]{ECF8E8}} \color[HTML]{000000} 0.19 & {\cellcolor[HTML]{F1FAEE}} \color[HTML]{000000} 0.13 & \bfseries {\cellcolor[HTML]{AEDEA7}} \color[HTML]{000000} 0.56 & {\cellcolor[HTML]{D1EDCB}} \color[HTML]{000000} 0.38 & {\cellcolor[HTML]{B6E2AF}} \color[HTML]{000000} 0.52 & {\cellcolor[HTML]{B4E1AD}} \color[HTML]{000000} 0.53 \\
Smoking & {\cellcolor[HTML]{F5FBF3}} \color[HTML]{000000} 0.09 & {\cellcolor[HTML]{F2FAEF}} \color[HTML]{000000} 0.13 & {\cellcolor[HTML]{EBF7E7}} \color[HTML]{000000} 0.20 & {\cellcolor[HTML]{F7FCF5}} \color[HTML]{000000} 0.07 & \bfseries {\cellcolor[HTML]{AEDEA7}} \color[HTML]{000000} 0.59 & {\cellcolor[HTML]{CDECC7}} \color[HTML]{000000} 0.42 & {\cellcolor[HTML]{B1E0AB}} \color[HTML]{000000} 0.57 & {\cellcolor[HTML]{B0DFAA}} \color[HTML]{000000} 0.58 \\
Income & {\cellcolor[HTML]{F4FBF1}} \color[HTML]{000000} 0.06 & {\cellcolor[HTML]{F1FAEE}} \color[HTML]{000000} 0.08 & {\cellcolor[HTML]{E9F7E5}} \color[HTML]{000000} 0.16 & {\cellcolor[HTML]{F7FCF5}} \color[HTML]{000000} 0.02 & {\cellcolor[HTML]{B2E0AC}} \color[HTML]{000000} 0.49 & {\cellcolor[HTML]{BBE4B4}} \color[HTML]{000000} 0.45 & {\cellcolor[HTML]{B0DFAA}} \color[HTML]{000000} 0.50 & \bfseries {\cellcolor[HTML]{AEDEA7}} \color[HTML]{000000} 0.51 \\
HomeValue & {\cellcolor[HTML]{EFF9EC}} \color[HTML]{000000} 0.11 & {\cellcolor[HTML]{E8F6E4}} \color[HTML]{000000} 0.18 & {\cellcolor[HTML]{E8F6E4}} \color[HTML]{000000} 0.18 & {\cellcolor[HTML]{F7FCF5}} \color[HTML]{000000} 0.02 & {\cellcolor[HTML]{B1E0AB}} \color[HTML]{000000} 0.52 & {\cellcolor[HTML]{C0E6B9}} \color[HTML]{000000} 0.45 & {\cellcolor[HTML]{B1E0AB}} \color[HTML]{000000} 0.52 & \bfseries {\cellcolor[HTML]{AEDEA7}} \color[HTML]{000000} 0.54 \\
NightLights & {\cellcolor[HTML]{F7FCF5}} \color[HTML]{000000} 0.28 & {\cellcolor[HTML]{EAF7E6}} \color[HTML]{000000} 0.40 & {\cellcolor[HTML]{DCF2D7}} \color[HTML]{000000} 0.49 & {\cellcolor[HTML]{EBF7E7}} \color[HTML]{000000} 0.39 & {\cellcolor[HTML]{BAE3B3}} \color[HTML]{000000} 0.66 & {\cellcolor[HTML]{C3E7BC}} \color[HTML]{000000} 0.62 & \bfseries {\cellcolor[HTML]{AEDEA7}} \color[HTML]{000000} 0.71 & \bfseries {\cellcolor[HTML]{AEDEA7}} \color[HTML]{000000} 0.71 \\
PopulationDensity & {\cellcolor[HTML]{F7FCF5}} \color[HTML]{000000} 0.33 & {\cellcolor[HTML]{EAF7E6}} \color[HTML]{000000} 0.41 & {\cellcolor[HTML]{E7F6E2}} \color[HTML]{000000} 0.43 & {\cellcolor[HTML]{F1FAEE}} \color[HTML]{000000} 0.37 & {\cellcolor[HTML]{BCE4B5}} \color[HTML]{000000} 0.58 & {\cellcolor[HTML]{DCF2D7}} \color[HTML]{000000} 0.47 & {\cellcolor[HTML]{B5E1AE}} \color[HTML]{000000} 0.60 & \bfseries {\cellcolor[HTML]{AEDEA7}} \color[HTML]{000000} 0.62 \\
TreeCover & {\cellcolor[HTML]{D5EFCF}} \color[HTML]{000000} 0.26 & {\cellcolor[HTML]{C9EAC2}} \color[HTML]{000000} 0.29 & {\cellcolor[HTML]{BEE5B8}} \color[HTML]{000000} 0.31 & {\cellcolor[HTML]{EBF7E7}} \color[HTML]{000000} 0.20 & {\cellcolor[HTML]{F7FCF5}} \color[HTML]{000000} 0.15 & {\cellcolor[HTML]{E5F5E1}} \color[HTML]{000000} 0.22 & {\cellcolor[HTML]{C3E7BC}} \color[HTML]{000000} 0.30 & \bfseries {\cellcolor[HTML]{AEDEA7}} \color[HTML]{000000} 0.34 \\
Elevation & \bfseries {\cellcolor[HTML]{AEDEA7}} \color[HTML]{000000} 0.49 & {\cellcolor[HTML]{B6E2AF}} \color[HTML]{000000} 0.45 & {\cellcolor[HTML]{DCF2D7}} \color[HTML]{000000} 0.25 & {\cellcolor[HTML]{D0EDCA}} \color[HTML]{000000} 0.32 & {\cellcolor[HTML]{F4FBF2}} \color[HTML]{000000} 0.05 & {\cellcolor[HTML]{F7FCF5}} \color[HTML]{000000} 0.02 & {\cellcolor[HTML]{D5EFCF}} \color[HTML]{000000} 0.29 & {\cellcolor[HTML]{C8E9C1}} \color[HTML]{000000} 0.37 \\
All metrics (mean) & {\cellcolor[HTML]{F4FBF1}} \color[HTML]{000000} 0.14 & {\cellcolor[HTML]{F2FAEF}} \color[HTML]{000000} 0.15 & {\cellcolor[HTML]{EEF8EA}} \color[HTML]{000000} 0.18 & {\cellcolor[HTML]{F7FCF5}} \color[HTML]{000000} 0.11 & {\cellcolor[HTML]{B8E3B2}} \color[HTML]{000000} 0.44 & {\cellcolor[HTML]{D1EDCB}} \color[HTML]{000000} 0.34 & {\cellcolor[HTML]{B4E1AD}} \color[HTML]{000000} 0.46 & \bfseries {\cellcolor[HTML]{AEDEA7}} \color[HTML]{000000} 0.48 \\
\end{tabular}
\end{table}

While the benefits of SatCLIP integration were evident in interpolation, a deeper look at the extrapolation results reveals a more complex narrative. Contrary to the interpolation scenario, the inclusion of SatCLIP features does not consistently lead to improved performance in extrapolation tasks. On an average PDFM without SatCLIP performs slightly better with an $R^2$ of 0.61, whereas the PDFM with SatCLIP scores an $R^2$ of 0.60.  In fact, 16 out of 27 tasks PDFM without SatCLIP, outperforms its SatCLIP-augmented counterpart. The other 11 tasks where the PDFM with SatCLIP does show improvement, the performance gains are only marginally better. 

\end{document}